\documentclass[conference]{IEEEtran}
\usepackage{times}

\usepackage[numbers]{natbib}
\usepackage{multicol}

\usepackage{booktabs}  
\usepackage{tabularx} 
\usepackage{amsmath} 
\DeclareMathAlphabet\mathbfcal{OMS}{cmsy}{b}{n}
\def\0{{\bf 0}}
\def\1{{\bf 1}}

\def\eg{\emph{e.g.}} 
\def\ie{\emph{i.e.}}

\usepackage{amsmath}

\usepackage{epigraph} 
\usepackage{todonotes}
\usepackage{dsfont}
\usepackage{booktabs}  
\usepackage{multirow}  
\usepackage{algorithm}

\usepackage{nicefrac}       %
\usepackage{xfrac} %
\usepackage{graphicx}     
\usepackage{amsmath}    
\usepackage{xspace}
\usepackage{subcaption}
\usepackage{amssymb}
\usepackage{amsfonts}
 \usepackage{bbding}

\usepackage{color}
\usepackage{colortbl}
\usepackage{xcolor}
\newcommand{\indicator}[1]{\mathds{1}{#1}} 
\definecolor{deemph}{gray}{0.6}
\newcommand{\gc}[1]{\textcolor{deemph}{#1}}
\definecolor{baselinecolor}{gray}{.9}
\definecolor{yellow}{RGB}{218,165,32}
\definecolor{lightcyan}{rgb}{0.88, 1.0, 1.0}
\definecolor{lightskyblue}{rgb}{0.53, 0.81, 0.98}
\definecolor{aliceblue}{rgb}{0.94, 0.97, 1.0}
\definecolor{LightSlateBlue}{RGB}{70,130,180}
\definecolor{DeepBlue}{RGB}{65,100,170}
\definecolor{DeepPurple}{RGB}{136,105,160}
\definecolor{LightGreen}{RGB}{59,125,35}
\definecolor{LightRed}{RGB}{234,66,53}
\definecolor{cvprblue}{rgb}{0.21,0.49,0.74}
\newcommand{\baseline}[1]{\cellcolor{aliceblue}{#1}}
\usepackage[pagebackref,breaklinks,colorlinks,citecolor=cvprblue,bookmarks=true]{hyperref}

\definecolor{taska}{RGB}{0,102,0}

\definecolor{taskb}{RGB}{0, 51, 102}

\definecolor{taskc}{RGB}{64,64,64}

\def\taska{\textcolor{taska}}
\def\taskb{\textcolor{taskb}}
\def\taskc{\textcolor{taskc}}

\usepackage{booktabs}

\usepackage{caption} %

\hypersetup{
 colorlinks=true,
 linkcolor=red,
 citecolor=cvprblue,
 filecolor=magenta, 
 }

\usepackage[capitalise]{cleveref}

\crefname{section}{Sec.}{Secs.}
\crefname{section}{Sec.}{Secs.}

\crefname{figure}{Fig.}{Figs.} %
\Crefname{figure}{Fig.}{Figs.} %

\crefname{table}{Table}{Tables} %
\Crefname{table}{Table}{Tables} %

\crefname{equation}{Eq.}{Eqs.}
\Crefname{equation}{Eq.}{Eqs.}

\newcommand{\modelname}{RISE\xspace}

\begin{document}

\title{
{\textit{\modelname}}:
Self-Improving Robot Policy with\\
Compositional World Model
}

\author{\authorblockN{
Jiazhi Yang$^{1,2*\dagger}$ \quad
Kunyang Lin$^{2*}$ \quad
Jinwei Li$^{2,6*}$ \quad
Wencong Zhang$^{2*}$ \quad
Tianwei Lin$^{5}$\quad
Longyan Wu$^{4}$ \\
Zhizhong Su$^{5}$ \quad
Hao Zhao$^{6}$ \quad
Ya-Qin Zhang$^{6}$ \quad
Li Chen$^{3}$ \quad
Ping Luo$^{3}$ \quad
Xiangyu Yue$^{1}$$^{\natural}$ \quad
Hongyang Li$^{3}$$^{\natural}$ \quad
\\
}
\vspace{0.1em}
\authorblockA{
$^{1}$ The Chinese University of Hong Kong  \quad
$^{2}$ Kinetix AI \quad
$^{3}$ The University of Hong Kong \\
$^{4}$ Shanghai Innovation Institute \quad
$^{5}$ Horizon Robotics \quad
$^{6}$ Tsinghua University \\
{\small{\gc{$^{*}$Equal Contribution}}} \quad
{\small{\gc{$^{\dagger}$Project lead}}} \quad 
{\small{\gc{$^{\natural}$Equal Advising}}} \\
\vspace{0.1em}
\texttt{\url{https://opendrivelab.com/RISE}}
}
}

\twocolumn[{%
	\renewcommand\twocolumn[1][]{#1}%

	\maketitle
        \vspace{-4mm}
	\begin{center}
        \includegraphics[width=1.0\textwidth]{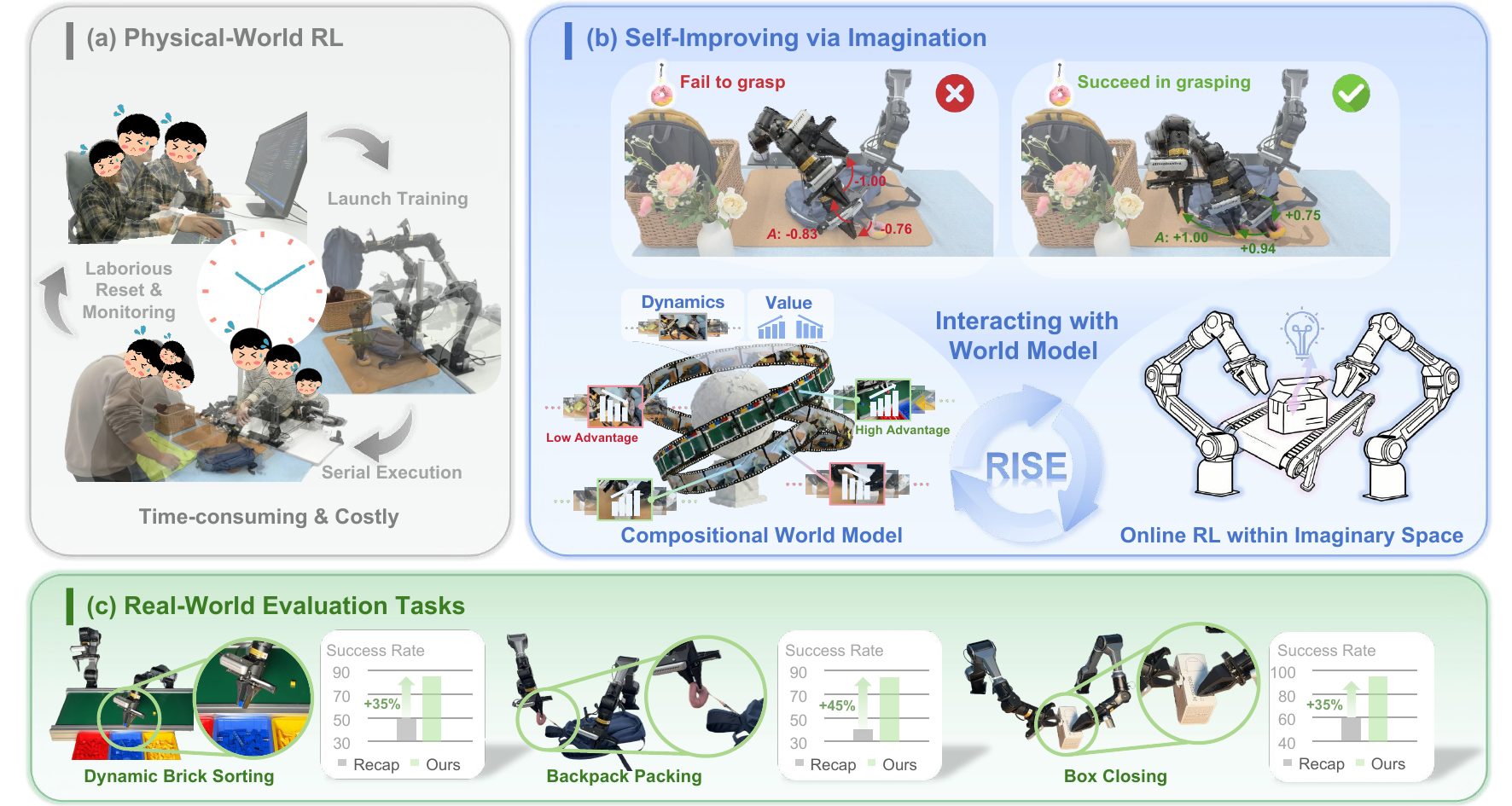}
		\captionof{figure}{We present \textbf{\modelname}, a framework for \textbf{R}einforcement learning via \textbf{I}magination for \textbf{SE}lf-improving robots.
        (a) 
        Conventional physical-world RL is bottlenecked by hardware cost, slow serial interaction, and the need for manual reset.
        (b) \modelname shifts the learning environment to a \textbf{Compositional World Model}, which first emulates future observations for proposed actions, then evaluates imagined states to derive advantage for policy improvement.
        (c)
        Training on massive imaginative rollouts effectively bootstraps \modelname's performance across a variety of complex, contact-rich tasks, surpassing prior art by a non-trivial margin.
        }
  \label{fig:teaser}
	\end{center}
}]

\pdfinfo{
   /Author (Homer Simpson)
   /Title  (Robots: Our new overlords)
   /CreationDate (D:20101201120000)
   /Subject (Robots)
   /Keywords (Robots;Overlords)
}

\begin{abstract}

Despite the sustained scaling 
on model capacity and data acquisition,
Vision–Language–Action (VLA) models
remain brittle in contact-rich and dynamic manipulation tasks, where minor execution deviations can compound 
into failures.
While reinforcement learning (RL) offers a principled path to robustness, 
on-policy RL in the physical world is constrained by safety risk, hardware cost, 
and 
environment reset.
To bridge this gap, we present \modelname,  
a scalable framework of 
robotic reinforcement learning via imagination.
At its core is a Compositional World Model that (i) predicts multi-view future via a controllable dynamics model, and (ii) evaluates imagined outcomes with a progress value model, producing informative advantages for the policy improvement.
Such compositional design allows state and value to be tailored by best-suited yet distinct architectures and objectives.
These components are integrated into a closed-loop self-improving pipeline that continuously generates imaginary rollouts, estimates advantages, and updates the policy in imaginary space without costly physical interaction.
Across three challenging real-world tasks, 
\modelname yields significant improvement over prior art,
with more than +35\% absolute performance increase in dynamic brick sorting, 
+45\% for backpack packing, and +35\% for box closing, respectively.
\end{abstract}

\IEEEpeerreviewmaketitle

\section{Introduction}

The trajectory of embodied intelligence has been reshaped by the scaling of foundation models.
Particularly,
VLA models have emerged as the dominant paradigm for generalist robot control, leveraging massive pre-training 
on web-scale data to acquire broad semantic understanding and instruction-following capabilities~\cite{brohan2022rt1, black2024pi0,intelligence2025pi05,kim2024openvla, Fang_RSS_24, shi2025diversity}.
Despite the progress on 
high-level semantic competence,
such VLAs still fall short of robust manipulation under complex physical dynamics, such as precise grasping of moving objects or effective bi-manual coordination~\cite{luo2025hil-serl, hu2025rac}.
This discrepancy highlights the 
inherent limitation
of Imitation Learning (IL), 
a core mechanism enabling VLAs to generate executable actions.
Concretely, IL is inherently limited by the quality and coverage of the expert demonstrations while suffering from the 
exposure bias
problem: once the robot drifts slightly off the expert's manifold, it lacks the recovery skills to correct its course, leading to compounding errors~\cite{ross2011dagger,kelly2019hgdagger,hu2025rac,chen2025_value_learning}.
Reinforcement Learning (RL), which improves agents through their own success and failure, offers a potential remedy.

In 
virtual simulators such as 
LIBERO~\cite{liu2023libero},
agents can play massive interactions in parallel, where both state and reward updates are controllable and accessible.
Such properties of highly-crafted simulators have inspired successful RL adaptations upon recent VLAs~\cite{lu2025vla,li2025simplevla,liu2025what_can}.
Nonetheless, such controllability and parallelization do not hold in a real-world regime, where 
robot executions are serial, time-consuming, and labor-intensive due to manual monitoring and resets,
as depicted in \Cref{fig:teaser}(a).
These physical challenges largely confine
previous methods of
real-world RL 
to offline data
with heavy distribution shift to current policy~\cite{xiao2025PLD, luo2024serl, luo2025hil-serl, wagenmaker2025dsrl}.
Ultimately, the policy improvement could be bottlenecked 
without sufficient on-policy data stream~\cite{levine2020offlineRLtutorial,yu2020mopo,AWR}.

The gap between the simulator and the physical world
motivates the development of world models, which first learn from passive experience and then simulate 
future outcomes conditioned on different actions~\cite{sutton_dyna,ha2018world_models, hafner2019dreamerv1, hafner2020dreamerv2, hafner2023dreamerv3,lecun2022path}.
Nevertheless, constructing a world model
applicable to real-world robotics
poses fundamental challenges.
For control, world models must faithfully follow actions to represent the accurate consequences.
Despite the improved visual realism by
integrating high-capacity generative models~\cite{yang2023unisim,guo2025ctrl,zhu2024irasim},
how to improve controllability over various actions remains an open problem~\cite{li2025worldmodelbench}.
Furthermore, 
learning from imagination necessitates
informative learning signals for intermediate actions, rather than relying solely on a binary 
indicator.
Otherwise, determining terminal success would require the world model to simulate the entire task execution, which is beyond the reliable horizon of most generative world models~\cite{li2025worldmodelbench,li2025comprehensive_survey}.

To handle these issues, we present \modelname, a 
holistic
learning framework 
that 
\textbf{R}einforces robot foundation model via \textbf{I}magination to enable \textbf{SE}lf-improving, as shown in \Cref{fig:teaser}(b).
At its core is an 
online learning environment
achieved by a learned world model.
Inspired by prior works that decompose 
world modeling into tractable sub-problems to flexibly leverage heterogeneous architectures and priors~\cite{barcellona2024dream,du2024video,zhou2024robodreamer,yang2025resim},
we 
build
a Compositional World Model that factorizes the simulation problem into two 
objectives, 
dynamics prediction and value estimation, 
allowing each to be instantiated with architectures and training objectives best suited to its role.

\begin{figure*}[ht!]
    \centering
		\includegraphics[width=1.0\textwidth]{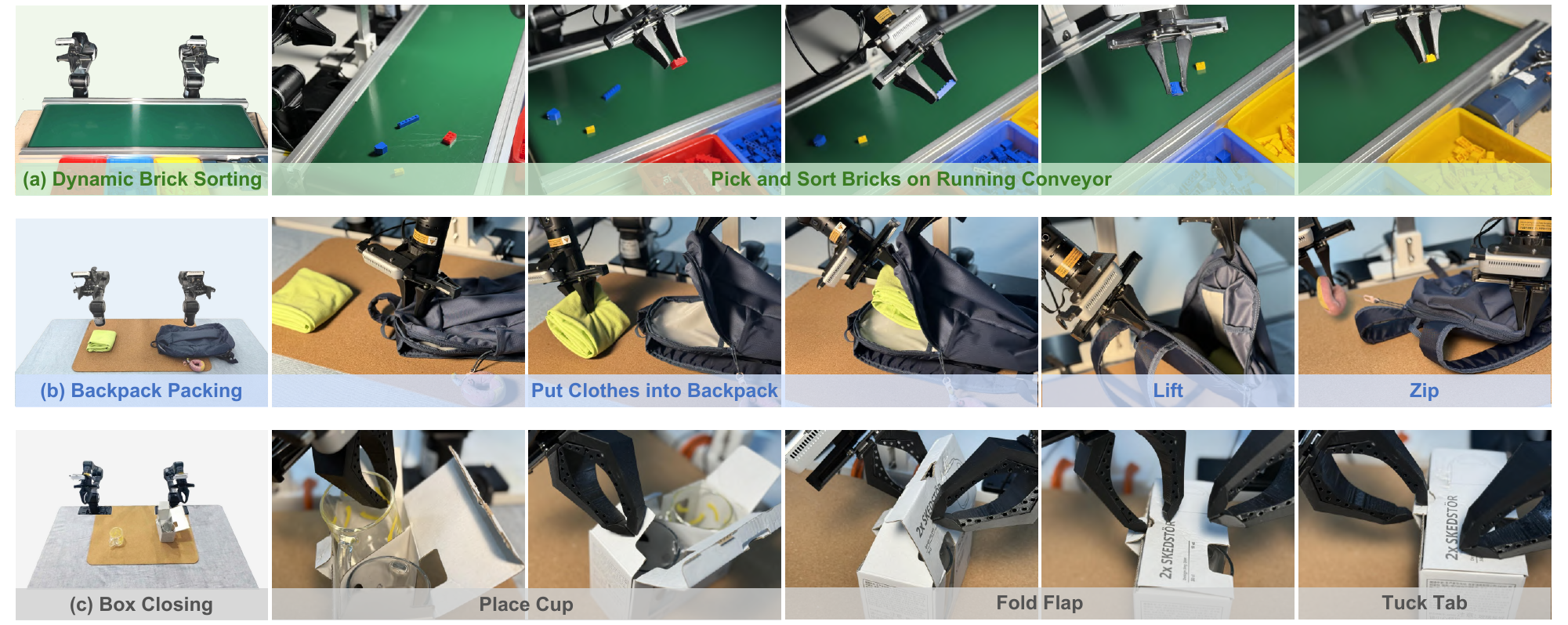}
		\caption{\textbf{Evaluation task suite of \modelname.} 
        \textbf{Left}: Tabletop setting.
        \textbf{Right}: Zoomed-in details of each task procedure.
        \textbf{\taska{Dynamic Brick Sorting}}
        involves 
        precisely picking up colored bricks from a moving conveyor and placing them into the corresponding color-designated bins.
        \textbf{\taskb{Backpack Packing}}
        requires the robot to open, insert clothes, lift, and zip the backpack.
        \textbf{\taskc{Box Closing}}
        necessitates subtle controls to fold the flap and tuck the tab into the box precisely.
        }
  \label{fig:task_sop}
\end{figure*}

Built on an efficient video diffusion model~\cite{liao2025genie, hacohen2024ltx}, we pre-train our dynamics model on large-scale robot datasets with a Task-centric Batching strategy to improve action controllability, which contributes to effective fine-tuning on targeted tasks.
The value model is initialized from a pre-trained VLA backbone~\cite{intelligence2025pi05} and adapted with both progress estimate~\cite{ma2024gvl,zhai2025vlac,ghasemipour2025self} and Temporal-Difference learning~\cite{Sutton_temporal_diff} objectives, providing dense and failure-sensitive evaluation of imagined states.
These components are combined to compute advantages for candidate actions, enabling stable policy improvement via advantage-conditioned training. 
As a result, \modelname performs on-policy reinforcement learning effectively in imagination.
As presented in  \Cref{fig:task_sop}, we rigorously evaluate 
\modelname on a suite of real-world tasks 
that stress-test dynamic adaptation and precision.
The results demonstrate that \modelname outperforms previous RL methods by a non-trivial margin, while avoiding costly real-world trial-and-error.

Our contributions are threefold:
\textbf{(1)} 
We propose \modelname, a principled framework for robotic reinforcement learning,
that enables autonomous self-improvement in a scalable and online manner.
\modelname
overcomes the physical restrictions posed by prior art 
by shifting the robotic interactions from physical environment to imaginative space. 
\textbf{(2)} 
At the core of this system is an online learning environment achieved by a Compositional World Model that
builds reliable dynamics and value estimates for real-world tasks.
We unveil 
critical design choices to derive stable learning signals for policy improvement.
\textbf{(3)}
Through extensive experiments on dexterous tasks, we demonstrate that \modelname 
exhibits significantly higher performance
compared to existing RL methods.

\textit{We view our work as the first 
study
on leveraging
world models as an effective learning environment 
for challenging real-world manipulation, bootstrapping performance on tasks requiring high dynamics, dexterity, and precision. Code is available at: \url{https://github.com/OpenDriveLab/RISE}.}

\section{Preliminary}

\subsection{World Model Formulation}

We aim to construct a world model consisting of a dynamics model for predicting future states and a value model for predicting rewards over different courses of action. Crucially, these predicted rewards are converted into advantages to guide RL training.
Formally, 
let $o_t = [m_t^1, \dots, m_t^n]$ be the multi-view observation at time $t$ with $n$ camera views. 
We apply a history window of length $N$ as $\mathbf{O}_t = \{o_{t-N}, \dots, o_{t-1}, o_t\}$ to capture temporal dependency.
The conditional action $\mathbf{a}_t$ is drawn from a running policy $\pi$ as $\mathbf{a}_t = [a_t, a_{t+1}, \dots, a_{t+H-1}] \sim \pi(\cdot | o_t, \ell)$,
where $\mathbf{a}_t$ is commonly applied as a sequence of actions with chunk length $H$, \ie, action chunk, and $\ell$ is a language instruction describing the task.
The dynamics model $\mathcal{D}$ predicts future observations $\{\hat{o}_{t+1}, \dots, \hat{o}_{t+H}\}$ conditioned on both the historical context and the proposed action sequence:
\begin{equation}
    \hat{o}_{t+1}, \dots, \hat{o}_{t+H} = \mathcal{D}(\mathbf{O}_t, \mathbf{a}_t).
\end{equation}

To evaluate the utility of imagined trajectories, we further introduce a value model 
$\mathcal{V}$, which assigns a progress signal towards successful completion conditioned on observation and task instruction as
$\mathcal{V}(\hat{o}_t, \ell)$.
We define the advantage as the average cumulative improvement across the entire chunk. Specifically, we compute the difference between the value of each predicted future observation $\hat{o}_{t+k}$ and the initial observation $o_t$ as the reward of action $a_{t+k}$, then take the expectation over the horizon of the action chunk as the advantage:
\begin{equation}
A(o_t, \mathbf{a}_t, \ell) = \left(\frac{1}{H} \sum_{k=1}^{H} \mathcal{V}(\hat{o}_{t+k}, \ell)\right)  - \mathcal{V}(o_t, \ell),
\label{eq:adv}
\end{equation}
where $A$ is associated with the action chunk proposed by the policy $\pi$, forming the learning signal for policy optimization.
The interaction between 
$\mathcal{D}$ and 
$\mathcal{V}$
occurs in imagination space, and both modules are compatible with \textit{multi-view} images.

\subsection{Reinforcement Learning} 
We formulate the problem as a standard RL setting with decision-making process as a Markov Decision Process (MDP) characterized by the tuple $(\mathcal{O}, \ell, \mathcal{A}, H, r)$. At each timestep $t$, given an observation $o_t \in \mathcal{O}$ and task instruction $\ell$, the policy $\pi$ generates an action sequence $\mathbf{a}_t \in \mathcal{A}^H$ of horizon $H$, obtaining reward $r$ for each step. The interaction between the policy and the environment induces a trajectory distribution $\rho_\pi(\tau)$, where $\tau = (o_0, \mathbf{a}_0, \dots, o_T) \in \mathcal{O} \times \mathcal{A} \cdots \mathcal{O}$. The objective is to maximize the expected return $\mathcal{J}(\pi) = \mathbb{E}_{\tau \sim \rho_\pi}[\sum_{t=0}^T r(o_t, \mathbf{a}_t)]$. To quantify the quality of a specific action sequence relative to the average policy performance, we utilize the advantage function $A^{\pi}(o_t, \mathbf{a}_t, \ell)$, estimated via \Cref{eq:adv}.

To ensure stable improvement over a reference policy $\pi_{\text{ref}}$, we adopt the probabilistic inference framework from $\pi^{*}_{0.6}$~\cite{amin2025pi06}. 
Rather than maximizing a regularized objective directly, we construct a target distribution $\hat{\pi}$ by weighting $\pi_{\text{ref}}$ with the probability of improvement $I$:
\begin{equation}
\hat{\pi}(\mathbf{a}_t|o_t,\ell) \propto \pi_{\text{ref}}(\mathbf{a}_t|o_t,\ell) \cdot p(I \mid A^{\pi_{\text{ref}}}(o_t, \mathbf{a}_t, \ell))^\beta.
\end{equation}
Since improvement is fully determined by the advantage value, we have $p(I|A^{\pi_{\text{ref}}}) \equiv p(I|\mathbf{a}_t, o_t, \ell)$. Applying Bayes' rule allows us to express the improvement likelihood as a density ratio:
\begin{equation}
p(I \mid \mathbf{a}_t, o_t, \ell) \propto \frac{\pi_{\text{ref}}(\mathbf{a}_t \mid I, o_t, \ell) p(I|o_t,\ell)}{\pi_{\text{ref}}(\mathbf{a}_t \mid o_t, \ell)}.
\label{eq:bayes_ratio}
\end{equation}
Substituting~\Cref{eq:bayes_ratio} into the target distribution and setting $\beta=1$ cancels the unconditional prior $\pi_{\text{ref}}$, yielding the simplified objective $\hat{\pi}(\mathbf{a}_t|o_t,\ell) = \pi_{\text{ref}}(\mathbf{a}_t \mid I, o_t, \ell)$. Practically, we implement this by conditioning the policy on discretized advantages, guiding generation toward high-return trajectories.

\section{Methodology}

Our approach is structured as follows: 
In \Cref{sec:world_model}, we propose a Compositional World Model that composes dynamics prediction with value estimation, providing an interactive environment with informative learning signals.
In \Cref{sec:policy_warmup}, we establish a Policy Warm-up stage on real-world experience to anchor the policy to practical behavioral distribution and equip it with advantage-conditioned capabilities.
In \Cref{sec:self_improving}, we present a Self-Improving Loop that iteratively generates imaginary rollouts and optimizes the policy within the world model.
Implementation details with compute allocation are covered in \Cref{sec:training_details}.

\subsection{
Compositional World Model 
}
\label{sec:world_model}

\begin{figure}[t]
    \centering
		\includegraphics[width=0.95\linewidth]{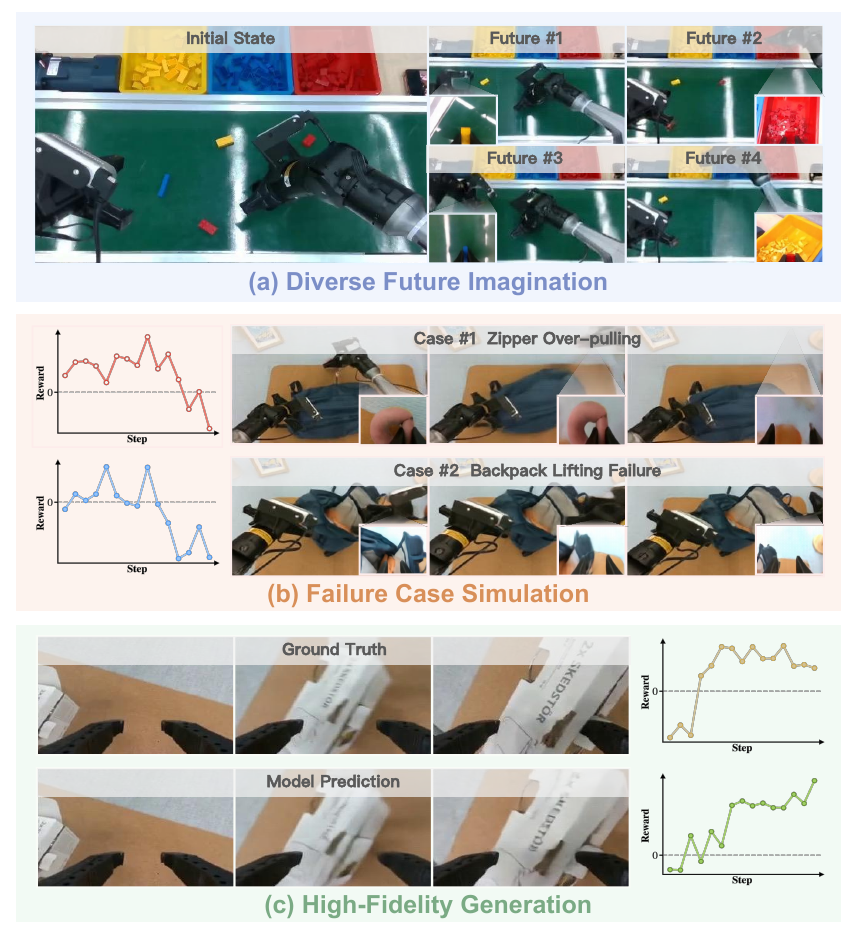}
		\captionof{figure}{\textbf{
        Qualitative imaginations produced by \modelname.
        }
        Given initial multi-view context and candidate action chunks, \modelname can (a) emulate a variety of future accordingly, (b) simulate failure cases with corresponding reward drops, and (c) maintain coherent predictions consistent with real executions.
        }
  \label{fig:dream}
\end{figure}

\begin{figure}[t!]
    \centering
    \includegraphics[width=1.0\linewidth]{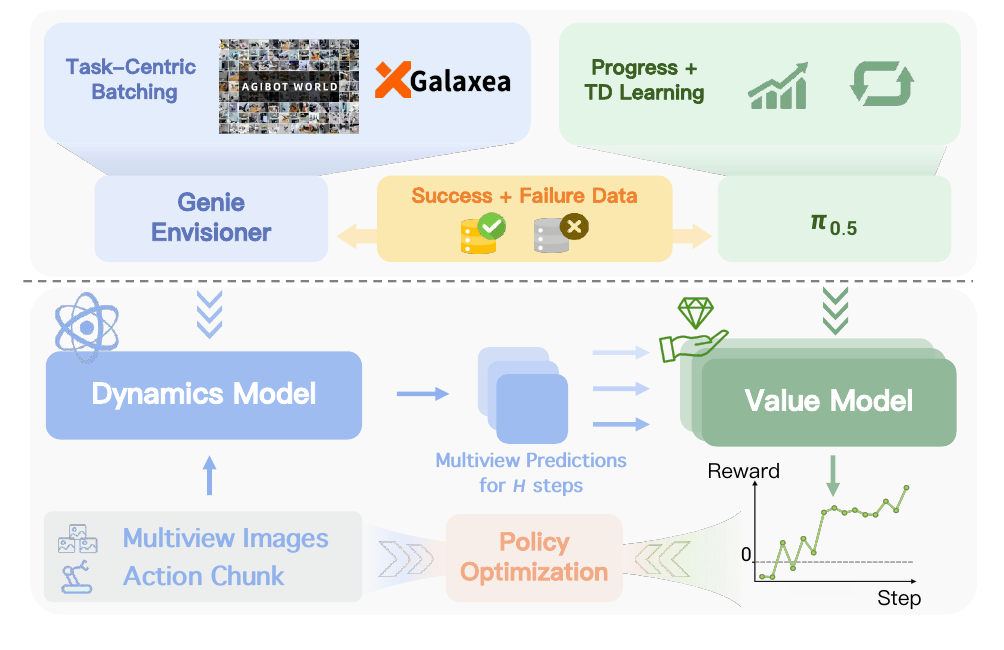}
    \caption{\textbf{
    Workflow of compositional world model.
    } 
    \textbf{Top}: Training recipe upon proper model initialization.
    \textbf{Bottom}: Inference pipeline that yields rewarded samples for policy optimization.
    Both modules are compatible with multi-view images.
    We omit text prompt for both policy and value model for brevity.
    }
    \label{fig:world_model}
\end{figure}

Scalable RL necessitates precise environment modeling to map current states and policy actions to future dynamics and rewards. To this end, we introduce a Compositional World Model to disentangle dynamics prediction from value estimation, thereby enabling independent architectural optimization for each component.
Starting from a context observation, the dynamics model emulates a faithful future under the candidate action chunk,
which would be evaluated by the value model to derive an advantage for policy improvement.
We show samples from imagination in \Cref{fig:dream} qualitatively.
Crucially, the model is employed exclusively during training, imposing zero computational overhead at inference.
The
training recipe and inference pipeline
of our world model are shown in \Cref{fig:world_model}.

\noindent \textbf{Controllable Dynamics Model.} 
Reliably simulating future states for RL yields two fundamental requirements:
(i) The generation latency should not be prohibitively high, which would bottleneck the throughput of the RL system.
(ii) The generated states should not only be plausible in visuals but also consistent with the conditional actions.
Thereby, we initialize our dynamics model from pre-trained Genie Envisioner~\cite{liao2025genie}, \ie, GE-base variant, 
which inherits the architectural advances in LTX-Video~\cite{hacohen2024ltx} and
features a favorable trade-off between generation quality and inference speed.
In comparison,
advanced world models such as Cosmos~\cite{ali2025cosmos2.5} takes more than 10 minutes for synthesizing 25 multi-view observations,
whereas GE only requires less than 2 seconds to achieve such a horizon, leading to 300x speedup.
Such generation efficiency is a critical pillar for applicable RL training.

Despite its efficiency, GE-Base is originally conditioned on text rather than fine-grained robot actions.
To endow the model with precise action controllability that could be further transferred into task-specific scenarios,
we further optimize the model on large-scale action-labeled datasets, including Agibot World~\cite{bu2025agibot} and Galaxea~\cite{jiang2025galaxea},
by incorporating an additional light-weight action encoder.
Additionally, we impose stronger noise on context frames compared to the original GE-base training, to improve the 
generation robustness when encountering motion blurs and visual artifacts that might occur in both recorded and synthesized data.
Nevertheless, fine-tuning a controllable world model on heterogeneous action data is prone to instability and slow convergence
when diverse tasks and visual domains are included within the same batch for each optimization iteration.
We mitigate this issue with a \textit{Task-Centric Batching} strategy, where each batch is sampled from a small fraction of tasks while covering more samples of the same task correlated with different actions. 
Intuitively, this batching strategy prioritizes
action diversity under the same scene over scenario diversity
for batch optimization,
thus contributing to improved action controllability.
Empirically, applying this strategy improves both task-specific fine-tuning efficiency, as in \Cref{tab:results_dynamics}, and stronger policy improvement, as in \Cref{tab:conveyor_ablation_module}.
With these design choices, our dynamics model is capable of providing fast and faithful multi-view state prediction to support the self-improving loop.

\noindent \textbf{Progress Value Model.}
Imagination-based policy improvement critically depends on a 
reward-related
signal that is (i) dense over long horizons and (ii) sensitive to subtle failures in contact-rich manipulation. 
We therefore learn a value estimator 
$\mathcal{V}$ that maps 
sensory
observations to a scalar value used to score imagined rollouts.
$\mathcal{V}$
is parameterized
from a pre-trained VLA policy $\pi_{0.5}$~\cite{intelligence2025pi05}, that brings in two advantages.
First, 
$\pi_{0.5}$ has been trained on broad robot datasets and thus carries robot-centric understanding that transfers naturally to value estimation.
Second, the policy backbone is compatible with multi-view inputs, whereas generic VLMs are mostly developed on single-view images without such adaptation.

As for training, we warm-start $\mathcal{V}$ with a simple temporal progress estimate as objective,
which equips our value model with a coarse understanding of monotonic temporal structure.
\begin{equation}
\mathcal{L}_{\text{prog}} = \mathbb{E}_{(o_t, \ell) \sim \mathcal{D}_{\text{exp}}} \left[ (\mathcal{V}(o_t, \ell) - \sfrac{t}{T})^2 \right],
\end{equation}
where $t$ indexes the current timestep within an episode of length $T$.
While progress regression provides a dense signal, it is often overly smooth and can be insensitive to failures, especially in contact-rich settings where execution errors might be subtle in visuals.
To conquer this, we augment the progress loss with Temporal-Difference (TD) learning~\cite{Sutton_temporal_diff}, which uses both successful demonstrations and failure rollouts to establish a value function that distinguishes success from errors.
\begin{equation}
\begin{aligned}
\mathcal{L}_{\text{TD}} 
&= \mathbb{E}_{(o_t, \ell, o_{t+1}) \sim \mathcal{D}}
\left[ (\mathcal{V}(o_t, \ell) - y_t)^2 \right], \\
y_t 
&= r_t + \gamma \mathcal{V}(o_{t+1}, \ell),
\end{aligned}
\end{equation}
where $\gamma$ is the temporal discount factor, and $r_t$ is set to 0 in intermediate steps while being $+1/-1$ at the end of successful and failure episodes, respectively.
Our final value learning objective simply combines both terms $\mathcal{L}_{\mathcal{V}} = \mathcal{L}_{\text{prog}} + \mathcal{L}_{\text{TD}}$ to leverage both the learning stability and error sensitivity provided by two terms, respectively.

\subsection{Policy Warm-up on Real-world Experience}
\label{sec:policy_warmup}

Before performing the on-policy improvement, we first warm-start the 
learning process 
with offline-collected data, 
which anchors the policy to a physically plausible behavior distribution on the targeted task, avoiding careless exploration in the later stage.
Both data composition and training objective mainly follow RECAP~\cite{amin2025pi06}.
For each task, we fine-tune the pre-trained policy, \ie, $\pi_{0.5}$~\cite{intelligence2025pi05},
on offline collected data, 
comprising 
expert demonstrations,
policy rollout with success and failure,
and human-intervened correction.
During training, the policy is conditioned on an advantage signal, labeled by our learned value model $\mathcal{V}$ 
as in \Cref{eq:adv}, 
by treating
$\hat{o}_{t+k}$ as later frames from an offline recorded video.
Different from the practice in RECAP that labels advantage for offline data and policy rollout, in early experiments,
we found that assigning advantages for both sources yields worse results than labeling for rollout only.
Thereby, only rollout data is assigned the learned advantages
whereas both expert and human correction data are directly paired with optimal advantages, denoted as $\indicator{}$,
in our experiments.
Consequently, this warm-up stage empowers the policy to absorb action data in different qualities, which is critical for the next
self-improvement 
stage that learns from trial-and-error in an online manner.

\begin{table*}[t!]
    \centering
    \caption{\textbf{Performance comparisons on real-world tasks.} We evaluate success rates and scores across three diverse tasks, ranging from dynamic sorting to precise packing. \modelname exhibits superior performance compared to baselines in all scenarios.}
    \label{tab:main_results}
    \small
    \setlength{\tabcolsep}{4mm}
    
    \scalebox{1.0}{
    \begin{tabular}{l|cc|cc|cc}
    \toprule
          \multirow{2}{*}{Method} & \multicolumn{2}{c|}{Dynamic Brick Sorting} & \multicolumn{2}{c|}{Backpack Packing} & \multicolumn{2}{c}{Box Closing} \\
         & Succ.\ (\%) & Score & {Succ.\ (\%)} & {Score} & {Succ.\ (\%)} & {Score} \\
    \midrule
    $\pi_{0.5}$~\cite{intelligence2025pi05}         & 35.00     & 8.28      & 30.00     & 4.25      & 35.00     & 7.50  \\
    $\pi_{0.5}$+DAgger~\cite{ross2011dagger,kelly2019hgdagger}  & 15.00     & 6.10      & 50.00     & 7.00      & 40.00         & 7.50     \\
    $\pi_{0.5}$+PPO~\cite{PPO}     & 10.00     & 7.68      & 35.00     & 5.88      & 10.00      & 4.75  \\
    $\pi_{0.5}$+DSRL~\cite{wagenmaker2025dsrl}    & 10.00     & 6.65      & 10.00     & 3.50      & 10.00      & 7.63  \\
    RECAP~\cite{amin2025pi06}  & 50.00     & 9.00      & 40.00     & 6.13      & 60.00     & 8.13  \\
    \midrule
     \baseline{\modelname(Ours)} & \baseline{\textbf{85.00}} & \baseline{\textbf{9.78}} & \baseline{\textbf{85.00}} & \baseline{\textbf{9.50}} & \baseline{\textbf{95.00}} & \baseline{\textbf{9.88}} \\
    \bottomrule
    \end{tabular}}
\end{table*}

\begin{figure}[t!]
    \centering
    \includegraphics[width=0.98\linewidth]{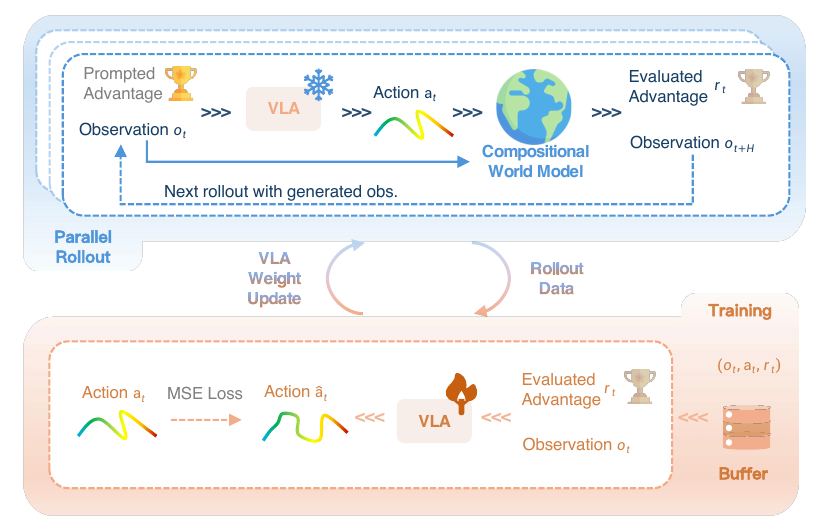}
    \caption{\textbf{
    Self-improving loop
    of \modelname.}
    Our learning pipeline encompasses two stages.
    \textbf{Top}: Rollout stage. 
    Prompted with an optimal advantage, the rollout policy interacts with the world model to produce rollout data.
    \textbf{Bottom}: Training stage. The behavior policy is then trained to 
    generate proper action under an advantage-conditioning scheme.
    } 
    \label{fig:rl_system}
\end{figure}

\subsection{
Self Improving
with World Model}
\label{sec:self_improving}
With the advantage conditioning capability acquired from
the warm-up stage on offline data,
we then apply the compositional world model as an interactive simulator to improve the policy.
The self-improving loop executes the Rollout stage and Training stage iteratively, as shown in~\Cref{fig:rl_system}.

\noindent \textbf{Rollout Stage.}
To start off,
we sample an initial state from the warm-up offline dataset.
Along with the observation, we additionally prompt 
the rollout policy $\pi_\text{rollout}$
with an optimal advantage $\indicator{}$,
to infer an action with positive intent.
\begin{equation}
\hat{\mathbf{a}}_t = \pi_{\text{rollout}}(\indicator{},o_t,\ell).
\end{equation}
Visual history and action proposal are
fed into the dynamics model to synthesize the next $H$ visual states. 
These imagined states are then evaluated by the value model to compute the actual advantage of the proposed action, denoted as $A^{\pi_{\text{rollout}}}(o_t,\hat{\mathbf{a}}_t,\ell)$. We define $\indicator{}$ as the prompted advantage for inferring optimal actions,
whereas 
$A^{\pi_{\text{rollout}}}(o_t,\hat{\mathbf{a}}_t,\ell)$ denotes the evaluated advantage, reflecting the true utility of the generated action.
This advantage is discretized into one of $N$ uniform bins representing the practical
advantage of the action in the current state.
To broaden the state coverage during the online training process,
imagined states would also serve as input for the subsequent rollout.
From each offline state, such consecutive interaction would be conducted at most two times, considering the known error accumulation issue of generative video models~\cite{huang2025self}.
The rollout policy parameters are updated via an Exponential Moving Average (EMA)~\cite{izmailov2018averaging}, blended from behavior policy weights.
One major difference between \modelname and prior approaches that also leverage world model as learning environment~\cite{jiang2025world4rl,zhang2025reinforcing,chandra2025diwa} is that \modelname avoids explicitly simulating terminal states to obtain rewards, yet produces chunk-wise advantage for proposed actions directly.

\noindent \textbf{Training Stage.}
The on-policy rollout data $\langle o, \hat{a}, A \rangle$ form batch samples to optimize the policy.
The VLA is trained to minimize the distance between its output and the proposed action $\hat{a}$, given the evaluated advantage $A$ as a condition. This allows the policy to learn from both high-advantage successes and low-advantage failures discovered in imagination.
To prevent catastrophic forgetting during exploration, we also mix offline labeled data into the batch data.
Both offline and online experiences are leveraged under
unified learning objective:
\begin{equation}
\pi(A^{\pi_{\text{rollout}}}(o,\hat{\mathbf{a}}_t,\ell),o_t,\ell) \to \hat{\mathbf{a}}.
\end{equation}
which is optimized under generic flow-matching criteria~\cite{black2024pi0,intelligence2025pi05}. 

\subsection{Implementation Details}
\label{sec:training_details}

\noindent \textbf{World Model Training.}
The dynamics model goes through two phases. The pre-training stage on Galaxea~\cite{jiang2025galaxea} and Agibot World~\cite{bu2025agibot} is conducted on 16 NVIDIA H100 
GPUs with a global batch size of 512, taking about seven days.
Subsequently, for task-specific fine-tuning, we utilize 8 NVIDIA H100 GPUs with a global batch size of 64, which takes about three days to complete.
Parameterized from a pre-trained VLA~\cite{intelligence2025pi05},
the value model is directly fine-tuned on task-specific data, thanks to the robot-centric knowledge inherited from the policy backbone.
We apply progress estimate loss only for the first 10k training steps and include TD learning loss additionally for the remaining 40k steps.
With a total batch size of 64 on 8 GPUs, the model converges in about one day of training.
Importantly, 
both modules of our world model are only applied during the policy learning phase,
thus posing \textbf{zero} inference overhead
during real-world policy execution.

\noindent \textbf{Policy Training.} 
The policy warm-up phase largely follows the training procedure of RECAP~\cite{amin2025pi06} on an offline collected dataset, where the policy is conditioned on advantage labeled by our learned value model.
The following self-improving stage then goes around 10k steps.
For both stages, global batch size is 64 on 8 GPUs.

\noindent \textbf{Task-specific Data.}
Both our world model and policy
share the same set of offline data for each task, including 
expert demonstrations and policy rollouts with success and failure, %
except that policy learning also consumes a fraction of DAgger data to enrich the recovery mode, similar to RECAP~\cite{amin2025pi06}.

\section{Evaluations}
We conduct a comprehensive evaluation to investigate the capabilities of \modelname. 
In particular, we focus on the
following questions:
\textbf{Comparative Analysis:} Does \modelname outperform existing mainstream RL and IL methods, particularly in real-world dexterous and long-horizon 
tasks?
\textbf{Design Choices}: How can the world model be effectively integrated into the RL loop, and is each module design essential?

\subsection{Real-world Experimental Setup}

Our real-world experiments employ a dual 7-DoF AgileX robot with absolute joint control. We benchmark three dexterous, long-horizon tasks,
including: \textbf{Dynamic Brick Sorting}: The robot is required to sort diverse bricks dynamically on an operating conveyor belt, shown in \Cref{fig:task_sop}(a), \textbf{Backpack Packing}: This task presents challenges involving compliant and deformable object manipulation as in \Cref{fig:task_sop}(b). \textbf{Box Closing}
The task requires precise bi-manual coordination to package a cup, as in \Cref{fig:task_sop}(c).
Notably,
ablations are conducted on the most challenging task in practice, \ie, Dynamic Brick Sorting. Hyperparameters remain fixed across variants.
Detailed robot setup and evaluation metrics are included in the Appendix. 

\subsection{Main Results}
\label{sec:main_results}
\noindent \textbf{Baselines.} We benchmark \modelname against state-of-the-art imitation and reinforcement learning baselines.
Each counterpart is developed with a close compute budget.
Implementation and data composition for each variant are detailed in the Appendix.
\begin{itemize}
    \item \textbf{$\pi_{0.5}$}~\cite{intelligence2025pi05}: A state-of-the-art VLA pre-trained on web-scale multi-robot data and fine-tuned on task demonstrations.
    \item \textbf{$\pi_{0.5}$ + DAgger}~\cite{ross2011dagger,kelly2019hgdagger}: An interactive baseline utilizing on-policy human corrections to mitigate exposure bias.
    
    \item \textbf{$\pi_{0.5}$ + PPO}~\cite{PPO}: A standard online RL baseline fine-tuning VLA weights via PPO.
    
    \item \textbf{$\pi_{0.5}$ + DSRL}~\cite{wagenmaker2025dsrl}: A sample-efficient method steering frozen VLAs by optimizing diffusion latent noise via RL.
    
    \item \textbf{RECAP}~\cite{amin2025pi06}: 
    An advantage-conditioned offline RL approach~\cite{frans2025cfgrl,kumar2019rewardcond} originally built off a proprietary pre-trained policy, \ie, $\pi_{0.6}$~\cite{amin2025pi06}.
    Due to the inaccessibility of $\pi_{0.6}$,
    we apply this approach to $\pi_{0.5}$ upon the same parameter-tuning and offline data corpora as ours.

\end{itemize}

\noindent \textbf{Results.} We present quantitative results in 
\Cref{tab:main_results}, reporting both \textit{Success Rate} and \textit{Stage-wise Score}, with evaluation criteria provided in the Appendix.
Although $\pi_{0.5}$ offers preliminary capability, we observe that online adaptation (PPO, DSRL) incurs severe instability. This leads to performance degradation, ~\eg, a sharp drop (35\%$ \to$ 10\%) in the \textit{Dynamic Brick Sorting} task.
 RECAP validates the benefit of advantage conditioning but falls short of \modelname. Notably, our method yields a 40\% margin in \textit{Backpack Packing}, while increasing success rates to 85\% and 95\% on the brick and box tasks, respectively.
Overall, \modelname significantly outperforms all RL and IL baselines across all tasks, with consistently high success rate.

\begin{table}[t!]
    \centering
    \vspace{4pt}
    \caption{\textbf{Ablation on offline data ratio.} Overall performance peaks at 0.6, indicating that balanced offline data is crucial for complex generalization.}
    \label{tab:abl_data_ratio}
    \small
    \setlength{\tabcolsep}{3.5mm}
    \scalebox{1.0}{
    \begin{tabular}{l|cccc}
    \toprule
          \multirow{2}{*}{Ratio} 
           & \multicolumn{1}{c}{Pick\&Place} & \multicolumn{1}{c}{Sort} & \multicolumn{2}{c}{Complete} \\
         & Succ.\ (\%) & Acc.\ (\%) & Succ.\ (\%) & Score \\
    \midrule
    0.1 & 15.00 & 83.33 & 5.00  & 1.35 \\
    0.3 & 78.75 & 80.95 & 25.00 & 7.03 \\
    \baseline{0.6} & \baseline{\textbf{90.00}} & \baseline{\textbf{87.50}} & \baseline{\textbf{50.00}} & \baseline{\textbf{8.32}} \\
    0.9 & 90.00 & 80.56 & 30.00 & 7.90 \\
    \bottomrule
    \end{tabular}}

\end{table}

\begin{table}[t!]
    \centering
    \caption{
    \textbf{Ablation on online action and state integration.} 
    Results demonstrate the necessity of 
    incorporating both online action proposed by the 
    rollout policy 
    and the online state generated by the 
    dynamics model. 
    }
    \label{tab:conveyor_ablation_new_state}
    \setlength{\tabcolsep}{1.5mm}
    
    \resizebox{\linewidth}{!}{%
        \begin{tabular}{ll|cccc}
            \toprule
            \multirow{2}{*}{Online Action} & \multirow{2}{*}{Online State} & \multicolumn{1}{c}{Pick\&Place} & \multicolumn{1}{c}{Sort} & \multicolumn{2}{c}{Complete} \\
                            &      & Succ.\ (\%)   & Acc.\ (\%)  & Succ.\ (\%) & Score            \\
            \midrule
            \XSolidBrush          & \XSolidBrush           & 80.00                       & 76.56                        & 35.00         & 6.98                \\
            \Checkmark            & \XSolidBrush           & 96.25                     & 84.42                    & 40.00     & 8.73             \\
            \baseline{\Checkmark} & \baseline{\Checkmark}  & \baseline{\textbf{98.75}} & \baseline{\textbf{92.41}}& \baseline{\textbf{70.00}} & \baseline{\textbf{9.43}} \\
            \bottomrule
        \end{tabular}%
    }
\end{table}

\begin{table}[t!]
    \centering
    \caption{
    \textbf{Ablations on the modular designs of dynamics and value models.} 
    ``w/o Progress'' indicates that the value model is trained without the auxiliary progress loss.
    Our full architecture proves to be the most effective across all metrics.
    }
    \label{tab:conveyor_ablation_module}
    \setlength{\tabcolsep}{1mm}

    \resizebox{\linewidth}{!}{%
        \begin{tabular}{ll|cccc}
            \toprule
            \multicolumn{2}{c|}{\multirow{2}{*}{Module Variants}}
                & \multicolumn{1}{c}{Pick\&Place}
                & \multicolumn{1}{c}{Sort}
                & \multicolumn{2}{c}{Complete} \\
            \multicolumn{2}{c|}{}
                & Succ.\ (\%)
                & Acc.\ (\%)
                & Succ.\ (\%)
                & Score \\
            \midrule

            \multirow{2}{*}{Dynamics}
                & w/o\ Pre-train
                & 97.50 & 60.26 & 15.00 & 7.43 \\
                & w/o\ Task-Centric
                & 93.75 & 89.33 & 40.00 & 8.78 \\
            \midrule

            \multirow{2}{*}{Value}
                & w/o\ Progress
                & 95.00 & 86.84 & 50.00 & 8.78 \\
                & w/o\ TD Learning
                & 98.75 & 72.15 & 35.00 & 8.38 \\
            \midrule

            \baseline{\modelname (Ours)}
                & \baseline{w/ all designs}
                & \baseline{\textbf{98.75}}
                & \baseline{\textbf{92.41}}
                & \baseline{\textbf{70.00}}
                & \baseline{\textbf{9.43}} \\
            \bottomrule
        \end{tabular}%
    }
\end{table}

\begin{figure}[t!]
    \centering
    \includegraphics[width=0.95\linewidth]{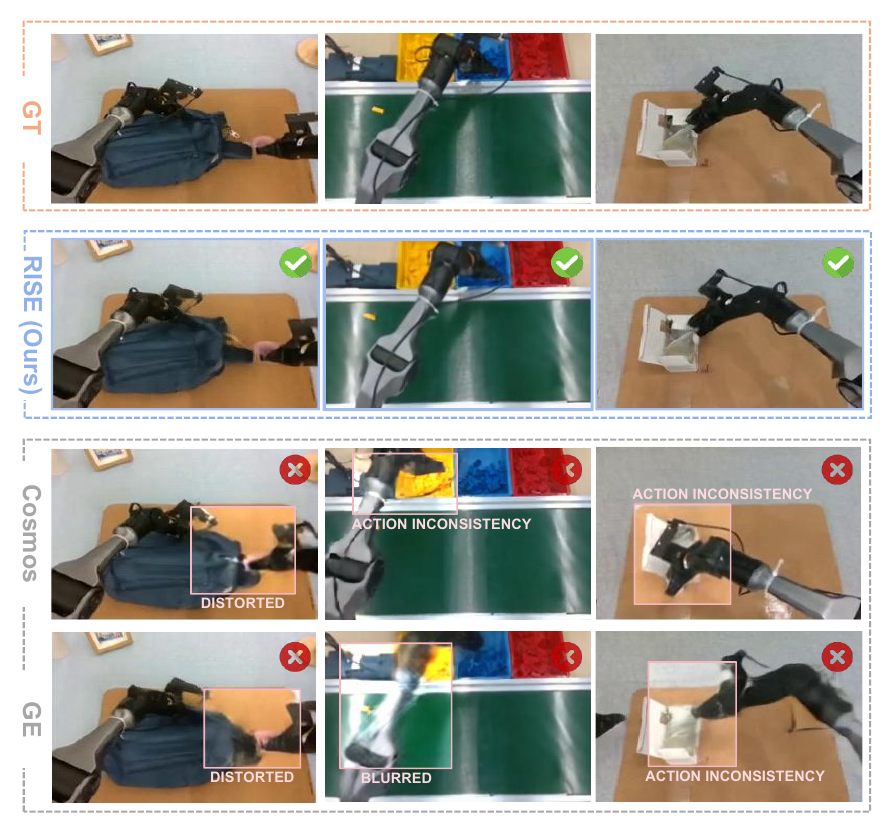}
    \caption{\textbf{Qualitative Comparison on Dynamics Models.}
    Cosmos~\cite{ali2025cosmos2.5} and Genie Envisioner~\cite{liao2025genie} suffer from geometric distortion, motion blurring, and physical inconsistency, whereas our method showcases temporally coherent and physically consistent results with Ground Truth (GT).}
    \label{fig:dynamics_case_by_case}
\end{figure}

\subsection{Ablation Study}
\label{sec:ablation}

\noindent \textbf{What ratio of the offline data should be allocated during RL training?}
Relying solely on online experience often leads to performance collapse due to the distribution shift between offline demonstrations and online rollouts. To address this, we investigate the optimal mixing ratio of offline data to retain performance. As shown in \Cref{tab:abl_data_ratio}, we observe a distinct trade-off. When the offline data ratio is too low (\eg, $0.1$), the success rate plummets to $5\%$. This confirms our hypothesis that insufficient offline retention leads to catastrophic forgetting in the face of massive online data. Conversely, an excessive ratio (\eg, $0.9$) also degrades performance. We attribute this to over-regularization, where the policy becomes too constrained to the offline distribution, hindering its ability to explore and discover superior policies.

\noindent \textbf{Can VLA models benefit from world-model generated online actions or states?} 
To validate this,
we evaluate three variants: a baseline without online signals, one with online actions only, and the full \modelname with both. Our results confirm the necessity of online signals. As shown in \Cref{tab:conveyor_ablation_new_state}, introducing online actions increases the success rate from $35\%$ to $40\%$. We attribute this improvement to the expanded action space exploration; unlike the static behavioral mode typically found in offline data, online rollouts allow the VLA to distinguish between high-advantage actions and suboptimal failures. Crucially, incorporating online states further raises the success rate to $70\%$. This suggests that dynamically generated online states provide a richer, virtually unbounded training distribution, overcoming the limitations of fixed offline datasets.

\begin{table}[t!]
    \centering
    \caption{\textbf{Quantitative comparison of dynamics models. }$\uparrow$ ($\downarrow$) denotes higher (lower) is better. Our method shows superior motion accuracy (EPE) and perceptual quality across both real-world tasks in \Cref{fig:task_sop} and the Bridge dataset~\cite{walke2023bridgedata}.
    }
    \label{tab:results_dynamics}
    \resizebox{\linewidth}{!}{
        \begin{tabular}{l ccccc}
            \toprule
            Method & PSNR $\uparrow$ & LPIPS $\downarrow$ & SSIM $\uparrow$ & FVD $\downarrow$ & EPE $\downarrow$ \\ 
            \midrule
            \multicolumn{6}{l}{\textit{Experiment \#1: Fine-tuning on our real world tasks}} \\
            \midrule
            Cosmos          & 21.17 & 0.14 & 0.79 & 97.90 & 1.21 \\
            GE         & 21.16 & 0.11 & 0.79 & 85.72 & 1.05 \\
            RISE (w/o Task-Centric) & 22.67 & 0.08 & 0.80 & \textbf{61.22}& 0.68 \\
            \baseline{\textbf{\modelname (Ours)}} & \baseline{\textbf{23.90}} & \baseline{\textbf{0.07}} & \baseline{\textbf{0.82}} & \baseline{66.84} & \baseline{\textbf{0.54}} \\
            \midrule
            \multicolumn{6}{l}{\textit{Experiment \#2: Fine-tuning on Bridge dataset~\cite{walke2023bridgedata}}} \\
            \midrule
            Cosmos          & 21.32 & 0.14 & 0.80 &  73.21 & 1.18 \\
            GE         & 21.47 & 0.12 & 0.79 &  64.55 & 0.96 \\
            RISE (w/o Task-Centric) & 22.61 & 0.10 & 0.78 & 49.07 & 0.72 \\
            \baseline{\textbf{\modelname (Ours)}} & \baseline{\textbf{23.68}} & \baseline{\textbf{0.10}} & \baseline{\textbf{0.82}} & \baseline{\textbf{45.21}} & \baseline{\textbf{0.64}} \\
            \bottomrule
        \end{tabular}
    }
\end{table}

\noindent \textbf{How significant is the impact of the modules on \modelname?} 
Quantitative results in~\Cref{tab:conveyor_ablation_module} highlight the criticality of each component. In the dynamics model, removing visual pre-training drops sorting accuracy by 32.15\% and completion to 15\%, underscoring the need for visual priors. Absence of task-centric design reduces completion by 30\%, validating the filtering of distractions. For the value model, ablating progress regression lowers success by 20\%, confirming the importance of dense signals. Furthermore, omitting TD learning leads to a 35\% decline, demonstrating its role in robust estimation.

\noindent\textbf{How reliable is the dynamics model?} 
We compare \modelname with Cosmos~\cite{ali2025cosmos2.5} and Genie Envisioner (GE)~\cite{liao2025genie} to investigate the reliability.
We evaluate generation quality using PSNR, SSIM~\cite{wang2004image}, LPIPS~\cite{zhang2018unreasonable}, and FVD~\cite{unterthiner2019fvd}, alongside optical flow end-point error (EPE)~\cite{zhang2025reinforcing} for action controllability. Quantitatively, \Cref{tab:results_dynamics} underscores the superiority of \modelname across all baselines under identical experimental settings. Notably, the significant reduction in EPE validates our task-centric pre-training, confirming that prioritizing action-conditioned dynamics effectively enhances motion awareness beyond standard pixel-level reconstruction. Qualitatively (\Cref{fig:dynamics_case_by_case}), while baselines suffer from blurring and kinematic inconsistencies, \modelname generates physically plausible dynamics with high fidelity. 
Additional comparisons are provided in the Appendix.

\section{Related Work}

\subsection{World Models for Robot Learning}
World models have been envisioned as a pathway to enable effective planning and learning through internal imagination~\cite{lecun2022path, ha2018world_models,hafner2019dreamerv1,sutton_dyna}. 
Early approaches in robotics and control 
focused on abstract state modeling in latent space with 
low-capacity dynamics model,
which are limited 
in 
capturing the rich visual and contact dynamics required for 
real-world manipulation~\cite{hafner2019dreamerv1,hafner2020dreamerv2,hafner2023dreamerv3,hansen2022tdmpc,hansen2023tdmpc2,hansen2022modem,lancaster2024modem_v2}.
Recent advances in large-scale generative modeling renewed world modeling in high-fidelity observation space~\cite{bruce2024genie,ali2025cosmos2.5,hafner2025dreamerv4,yang2023unisim,yang2024video,zhu2024irasim}.
However, adapting such models to serve as interactive environments for reinforcement learning remains challenging. 
Most approaches prioritize visual plausibility over action controllability, incurring prohibitive inference costs that prevent their use inside a reinforcement learning loop.
Beyond dynamics prediction, reward and value shaping also introduce an additional bottleneck to apply these models to policy improvement. 
Prior efforts heavily rely on sparse terminal rewards
or heuristic distance towards the goal state, which provide insufficient guidance for long-horizon manipulation and are brittle under long-term prediction errors~\cite{zhou2024dinowm,zhang2025reinforcing,zhu2025wmpo,mendonca2023structured}. 
Importantly, prior works center around either simulated benchmarks~\cite{hansen2022tdmpc,hansen2023tdmpc2,hafner2019dreamerv1,hafner2020dreamerv2,hafner2023dreamerv3,hafner2025dreamerv4,hansen2022modem, gao2025adaworld},
low-level control problems~\cite{li2025robotic,wu2023daydreamer,hansen2024hierarchical,roth2025learned}, 
or short-term tasks (\eg, pick and place), with limited validation in real-world tasks under contact-rich and complex dynamics~\cite{zhang2025reinforcing,chandra2025diwa,jiang2025world4rl,hung2025nora,lancaster2024modem_v2,guo2025ctrl,jang2025dreamgen,assran2025vjepa2,zhu2025uwm}.
Motivated by prior efforts that carefully integrate heterogeneous modules to tackle the challenging world modeling problem~\cite{zhou2024robodreamer,barcellona2024dream,du2024video},
we seamlessly compose a dynamics model and a value function
to achieve faithful simulation for various actions.

\subsection{Reinforcement Learning for Foundation Policies}

Reinforcement learning is increasingly used to strengthen VLA foundation policies on robustness and precision of manipulation.
A large body of work adapts VLA post-training with RL within simulated environments~\cite{liu2023libero,mu2025robotwin,chen2025robotwin2,mees2022calvin}, where interactions are cheap, resettable, and parallelizable~\cite{lu2025vla,li2025simplevla,chen2025pi_,liu2025what_can,chen2025tgrpo}.
However, such scalability does not hold in the physical world, where interactions are serial, slow, and labor-intensive.
Thereby, prior work on real-world RL is constrained to 
heavily reuse 
off-policy data
while online interactions are performed on limited robot hardware only, which potentially bottlenecks the policy improvement and is hard to scale~\cite{luo2025hil-serl,xiao2025PLD,RLPD,luo2024serl}.
Regarding learning stability, some work proposes to freeze the large-scale pre-trained policy while optimizing an additional residual policy~\cite{xiao2025PLD} or input noise distribution only~\cite{wagenmaker2025dsrl, li2025gr_rl}.
With most parameters unchanged, such approaches sacrifice the adaptability of the policy to target tasks.
In contrast, 
RECAP~\cite{amin2025pi06} enables finetuning the pre-trained policy via an advantage-conditioned formulation
~\cite{frans2025cfgrl, kumar2019rewardcond}, 
eliminating the complexity of adjusting the denoising chain for diffusion or flow-matching policy~\cite{lei2025rl}.
To derive reliable advantages for policy optimization, 
recent works resort to vision language models with
a progress estimate formulation, which is numerically 
stable and free from laborious annotations~\cite{ma2024gvl, zhang2025rewind,amin2025pi06,zhai2025vlac,ghasemipour2025self}.
However, such an objective is prone to the over-fitting problem and is less sensitive to subtle failures~\cite{amin2025pi06,li2025gr_rl}.
Distinguished from prior approaches,
we enable on-policy RL by shifting the learning 
environment from the physical world into an imaginative space via a learned world model.
Furthermore, our value 
model benefits from both progress estimate and 
Temporal-Difference learning~\cite{Sutton_temporal_diff} in stability and 
failure sensitivity.

\section{Conclusion} 
\label{sec:conclusion}

We introduced \modelname, a framework for on-policy reinforcement learning of robot foundation policies through imagination. \modelname replaces the physical environment with imagination during training, enabling scalable online improvement without the prohibitive cost and risk of real-world exploration.
Central to the system is a compositional world model that 
coherently orchestrates dynamics and value models,
built from proper recipes, 
to
efficiently emulate state and estimate advantage
for policy improvement.
Across real-world tasks spanning dynamic interaction, deformable-object handling, and bi-manual coordination, \modelname consistently outperforms strong post-training baselines,
proving that
world models can be applied as an effective learning environment to improve policy performance on challenging manipulation tasks. We hope this work serves as a reference for the community in exploring scalable self-improving VLA models.

\vspace{2mm}
\section{Limitations and future work} 
\label{sec:limit}

\noindent \textbf{The Gap between Imagination and Realism.} 
The effectiveness of \modelname is constrained by the accuracy and coverage of the learned world model. Although our compositional design improves controllability and consistency relative to prior generative simulators, the model can still produce physically implausible transitions in rare or underrepresented scenarios.
Addressing this gap requires future work on uncertainty-aware imagination and
principled integration of physical constraints
that explicitly encode geometry properties.

\noindent \textbf{The Simulated–Real Data Balance.}
Our results indicate that a non-trivial amount of real-world data remains essential to anchor the learning procedure. 
However, the optimal ratio between simulated rollouts and real-world experience requires further parameter tuning. 
Understanding the effectiveness and principles of these offline data represents an open problem.

\noindent \textbf{From Physical Cost to Compute Cost.}
\modelname shifts the primary bottleneck in robot learning from physical interaction to computation. 
While this trade-off releases the burden of physical interaction,
training high-fidelity world models incurs a high computational cost.
Improving the efficiency of world models will be critical for the compute-constrained regime.

\noindent \textbf{Outlook.}
Taken together, 
these limitations suggest a promising pathway in integrating learned simulation into a broader data ecosystem, where model-based reinforcement learning 
complements 
scarce
physical interaction. Discovering the right balance between these two key components points to a future of adaptive, robust, and sample-efficient robotic intelligence.

\section{Acknowledgments}
This study is supported by National Natural Science Foundation of China (62206172). This work is in part supported by the JC STEM Lab of Autonomous Intelligent Systems funded by The Hong Kong Jockey Club Charities Trust.

{
\bibliographystyle{plainnat}
\bibliography{bibliography_short, references}
}

\clearpage
\newpage

\section*{Appendix}

The appendix is organized as follows:

\begin{itemize}
    \item In \textbf{Appendix~\Cref{sec:additional_results}}, we present additional experimental results, including some ablation studies on minor components and visualization of the learned representations.
    \item In \textbf{Appendix~\Cref{sec:appendix_real_world_experiment_details}}, we list the comprehensive experimental settings, including hyperparameter configurations and hardware environments.
    \item In \textbf{Appendix~\Cref{sec:implementation_details}}, we elaborate on the implementation details of modules of \modelname.
    \item In \textbf{Appendix~\Cref{sec:conceptual_comparisons_related_work}}, we provide a conceptual comparison between our method and several highly related works.
    \item In \textbf{Appendix~\Cref{sec:qualitative_visualizations}}, we provide more qualitative visualizations.
    \item In \textbf{Appendix~\Cref{sec:failure_modes}}, we analyze failure mode of \modelname.
    \item In \textbf{Appendix~\Cref{sec:ref_vla}}, we include additional related work on vision-language-action models.
    \item In \textbf{Appendix~\Cref{sec:license}}, we list the license for each asset used in this paper, \ie, data and pre-trained weights.
    \item In \textbf{Appendix~\Cref{sec:impact}}, we envision the broader impact of the proposed method.
\end{itemize}

\section{Additional Results}
\label{sec:additional_results}

\begin{figure}[h!]
    \centering
    \includegraphics[width=1.0\linewidth]{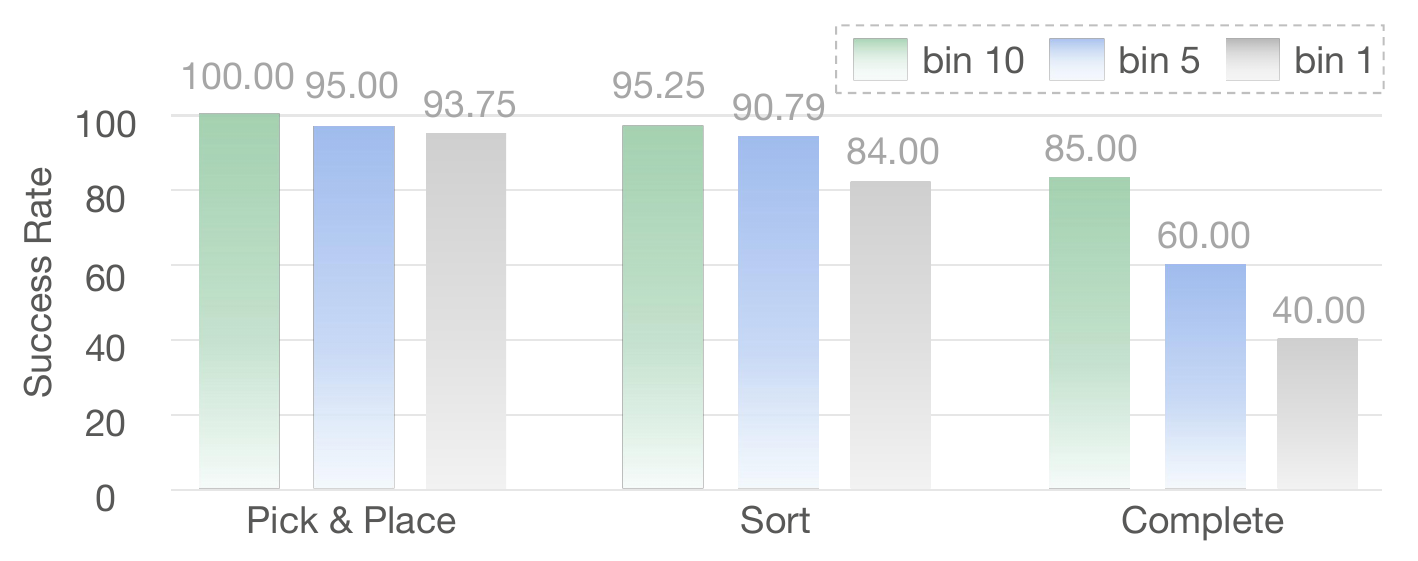}
    \caption{\textbf{Task success rate across advantage bins.} A clear performance drop is observed from high to low advantage levels, especially in Sorting. This confirms that our policy effectively captures behavior diversity through advantage conditioning.}
    \label{fig:bin_bar}
\end{figure}

\noindent \textbf{Can bins with different advantages reveal different performance of the policy?} \modelname utilizes advantage-based bins to guide RL training. We investigate whether the policy yields diverse task performance when conditioned on different bins. 
To this end, we evaluate the policy conditioned on high (Bin 10), neutral (Bin 5), and low (Bin 1) advantage bins.
Results in \Cref{fig:bin_bar} show a performance drop from bin 10 to bin 1, which supports our hypothesis. This performance drop is primarily attributed to sorting errors, as the success rate for sort deteriorates more significantly than for pick-and-place. Furthermore, the agent displays increased instability with lower bin indices. 
These findings demonstrate that our learned advantages are convincing and that the policy effectively captures the diversity of behaviors through our conditioning RL.
\begin{figure}[t!]
    \centering
    \includegraphics[width=1.0\linewidth]{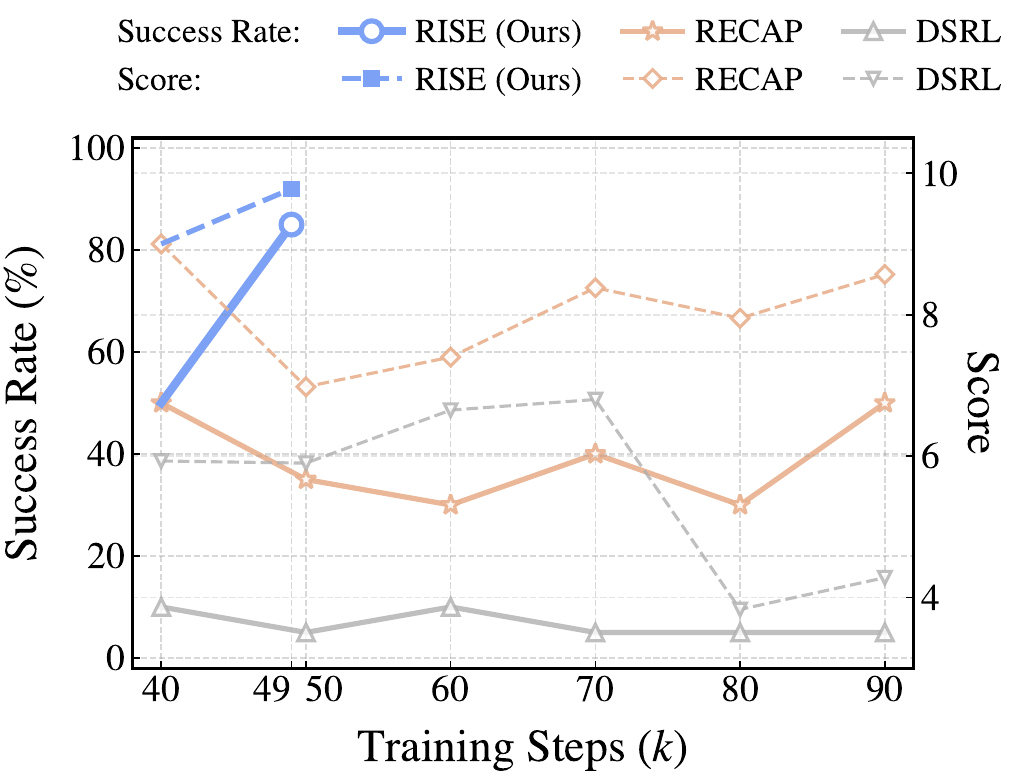}
    \caption{\textbf{
    Learning dynamics of
    RL alternatives.
    } 
    Compared to RECAP~\cite{amin2025pi06} and DSRL~\cite{wagenmaker2025dsrl},
    \modelname yields significantly higher results, 
    which cannot be attained by the competing methods even with extended training~\cite{amin2025pi06} and increased real-world interactions~\cite{wagenmaker2025dsrl}.
    } 
    \label{fig:compare_curve}
\end{figure}

\noindent \textbf{Can extended training of RL baselines match \modelname's performance?} 
To verify that our gains are not simply due to more training,
we extended the RECAP and DSRL baselines with an extra 50k steps under the same batch size of our method.
As shown in \Cref{fig:compare_curve}, 
RECAP saturates at a $30\%$ to $50\%$ success rate, while DSRL saturates at $5\%$ to $10\%$. In contrast, \modelname yields a $+35\%$ improvement (boosting success rate from $50\%$ to $85\%$) with only $9k$ additional steps. We attribute this efficiency to online world model interaction, providing diverse samples to mitigate overfitting.

\begin{figure}[t!]
    \centering
    \includegraphics[width=1.\linewidth]{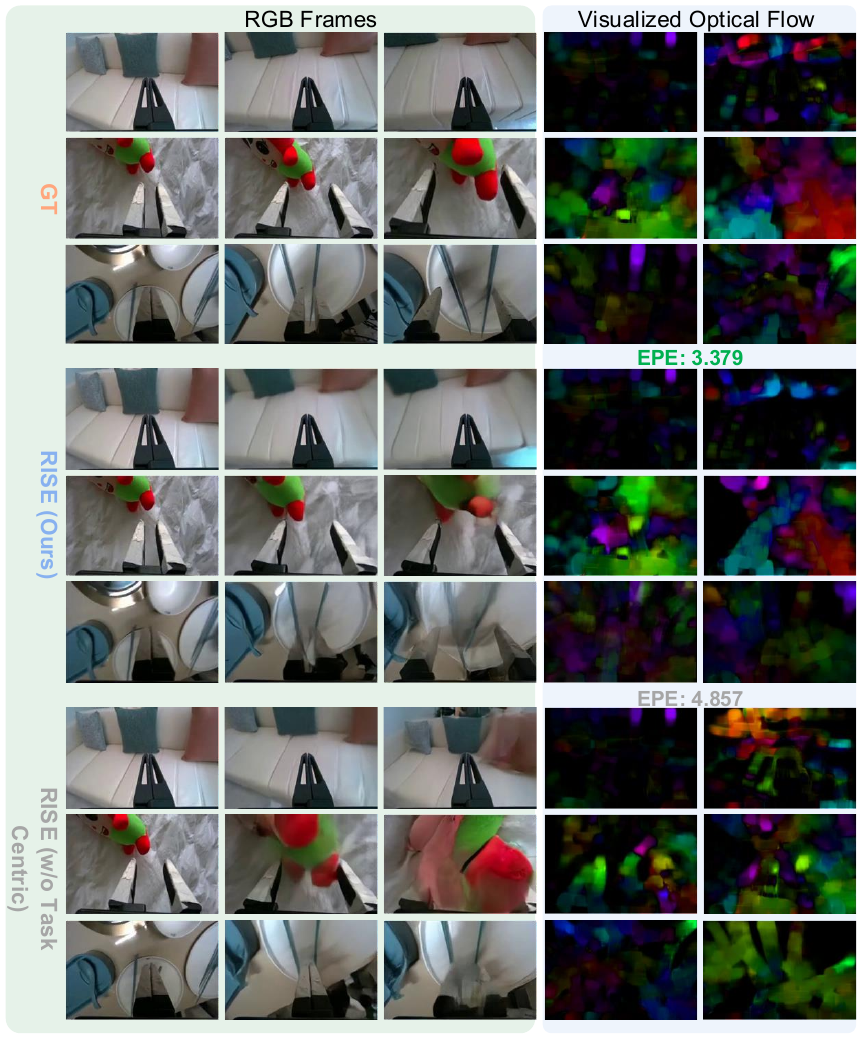}
    \captionof{figure}{\textbf{
    Task-centric versus non-task-centric during pre-train stage.} The optical flow maps demonstrate that our method captures action adherence more effectively during the initial stages of pre-training.}
  \label{fig:dynamics_during_pre-train_task_centric}
\end{figure}

\begin{figure*}[t!]
    \centering
    \includegraphics[width=1.0\textwidth]{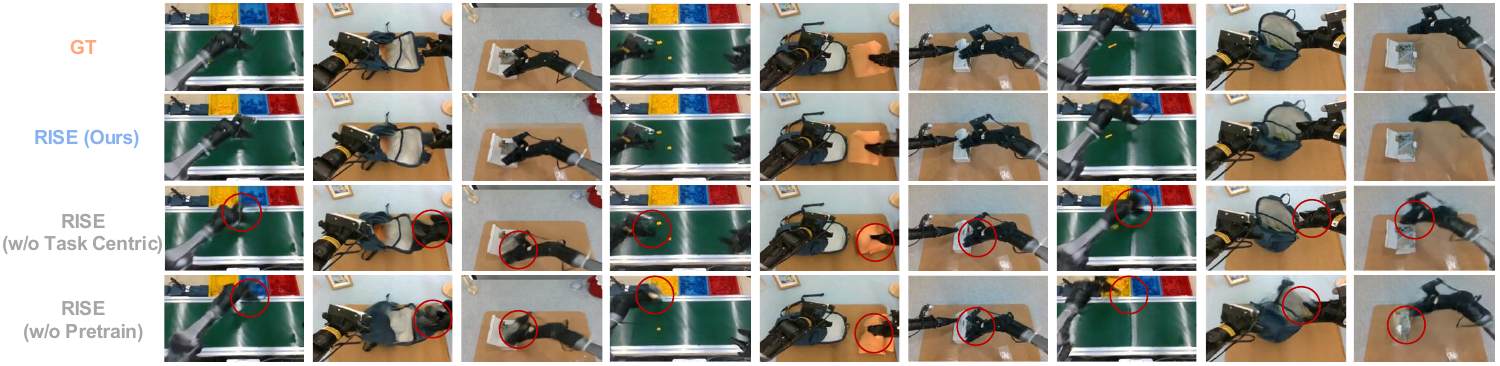}
    \captionof{figure}{\textbf{Visual ablation study on training strategies.} Compared to the other baselines, which exhibit significant degradation in image quality and motion coherence, our proposed method generates sharper, physically consistent predictions that strictly adhere to control actions.}
  \label{fig:dynamics_task_centric}
\end{figure*}

\noindent \textbf{What is the impact of pre-training and task-centric strategies on the generation quality of future dynamics?} 
We investigate the impact of strategies on the generation quality of future dynamics. As shown in ~\Cref{tab:dynamics_pre-train}, pre-training significantly enhances video generation fidelity. Moreover, ~\Cref{fig:dynamics_task_centric} provides visual comparisons, revealing that ablated variants (specifically \textit{w/o task-centric} and \textit{w/o pre-train}) suffer from action misalignment and severe blurring, whereas our method maintains high consistency with ground truth dynamics. Additionally, a sample-wise optical flow analysis in~\Cref{fig:dynamics_during_pre-train_task_centric} isolates the role of the task-centric mechanism. The results demonstrate that this objective effectively enhances motion sensitivity, yielding sharper and more physically coherent predictions.

\begin{table}[t]
    \centering
    \caption{
    \textbf{Quantitative ablation on the pre-training of our dynamics model.}
    }
    \label{tab:dynamics_pre-train}
    \resizebox{\linewidth}{!}{
    \begin{tabular}{@{}lccccc@{}}
        \toprule
        Method & PSNR $\uparrow$ & LPIPS $\downarrow$ & SSIM $\uparrow$ & FVD $\downarrow$ & EPE $\downarrow$ \\
        \midrule
        RISE (w/o pre-train) & 20.95 & 0.11 & 0.78 & 83.36 & 1.09 \\
        \baseline{\textbf{\modelname (Ours)}} & \baseline{\textbf{23.90}} & \baseline{\textbf{0.07}} & \baseline{\textbf{0.82}} & \baseline{\textbf{66.84}} & \baseline{\textbf{0.54}} \\
        \bottomrule
    \end{tabular}
    }
\end{table}

\begin{figure*}[t!]
    \centering
    \includegraphics[width=1.0\textwidth]{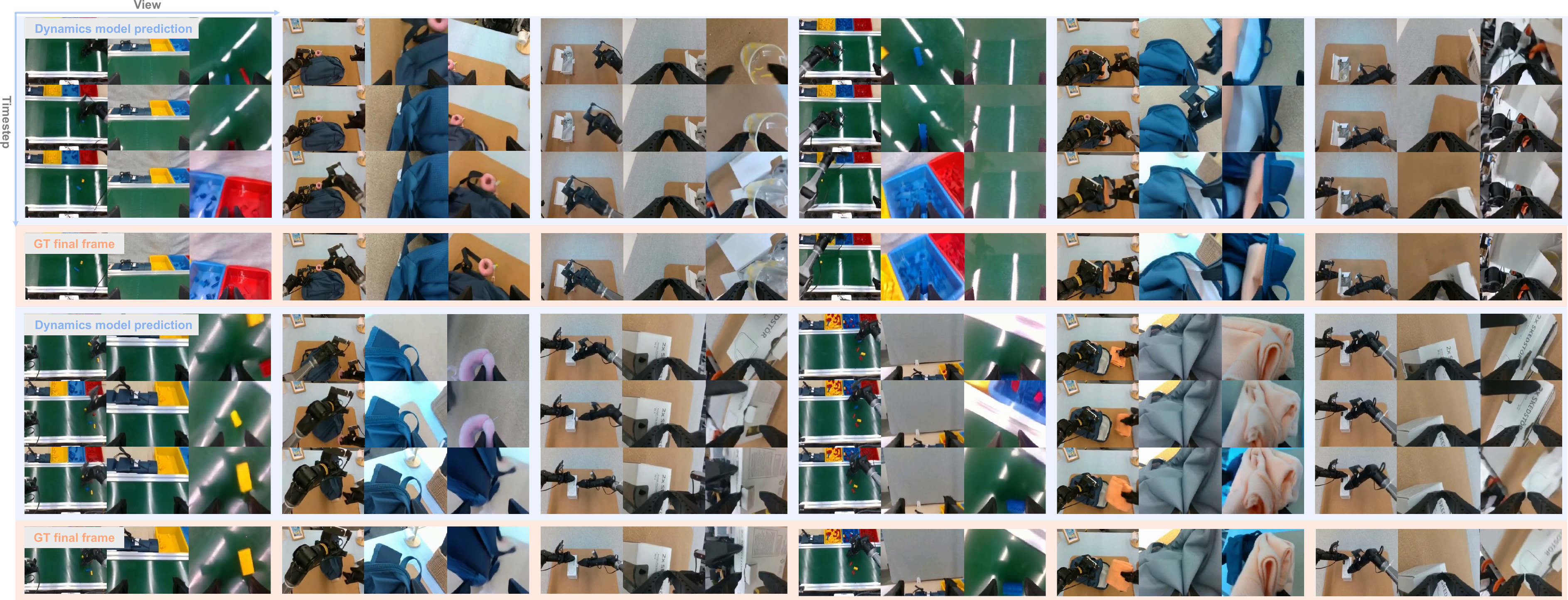}
    \captionof{figure}{\textbf{More multi-view rollouts on real world tasks.} Our dynamics model synthesizes coherent multi-view video rollouts with high visual fidelity, laying a solid foundation for reinforcement learning. Each video clip is ordered top to bottom.}
  \label{fig:dynamics_task_multi_view_rollout}
\end{figure*}

\section{Real-world Experimental Details}
\label{sec:appendix_real_world_experiment_details}

\subsection{Task Evaluation Standard}
\label{sec:task_details}

\begin{figure*}[t!]
    \centering
    \includegraphics[width=1.0\textwidth]{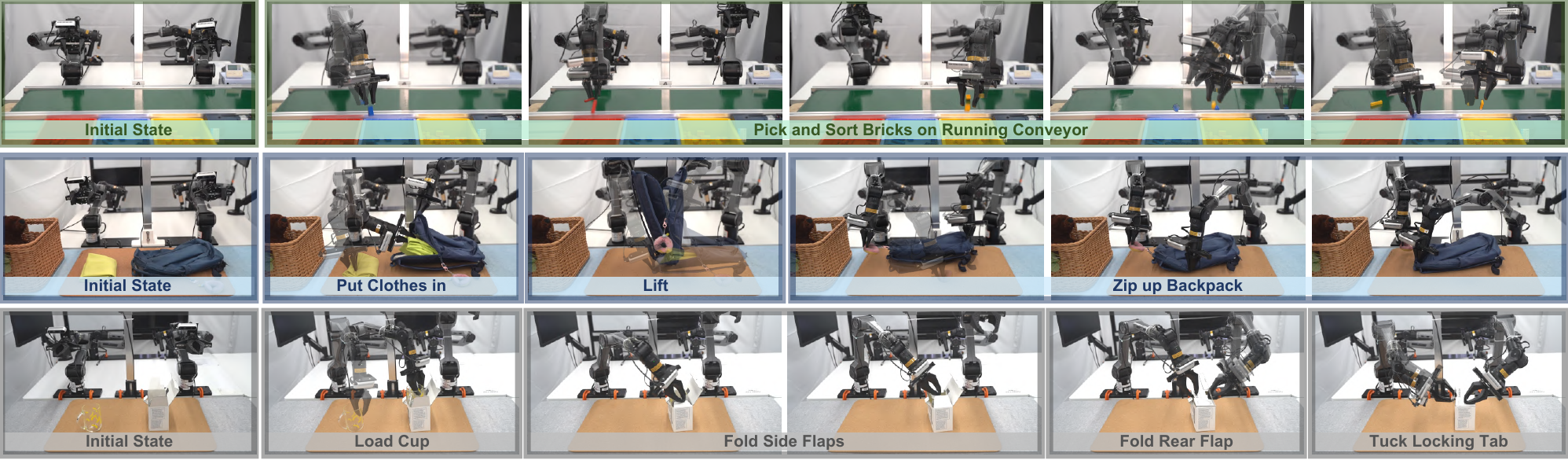}
    \caption{\textbf{Policy rollout.} \modelname demonstrates robust performance across diverse manipulation regimes. \textbf{\taska{Top:}} Handling dynamic scenes by sorting bricks on a moving conveyor. \textbf{\taskb{Middle:}} Manipulating deformable objects in the Backpack Packing task. \textbf{\taskc{Bottom:}} Achieving high-precision bi-manual control in Box Closing.}
    \label{fig:policy_rollout}
\end{figure*}

To provide a fine-grained analysis of policy performance beyond binary success, we define a quantitative evaluation rubric detailed in Table~\ref{tab:task_eval}. 
Given that our tasks involve multi-stage and long-horizon planning (as qualitatively illustrated in Figure~\ref{fig:policy_rollout}), a simple success/failure metric fails to capture the incremental progress of the agent. Therefore, we decompose each task into distinct sub-goals, with a total score of 10 per episode to ensure consistency across different tasks.
Throughout the paper, each of our evaluation results is based on an average score of 20 autonomous trials.

For the \textbf{Dynamic Brick Sorting} task, the scoring mechanism focuses on both manipulation robustness and classification accuracy. As the task involves processing a stream of objects, points are accumulated dynamically: successful grasping rewards the robot's low-level control, while correct placement into color-matched bins rewards semantic understanding. The score is capped at 10 to represent perfect clearing of the workspace.

For the \textbf{Backpack Packing} and \textbf{Box Closing} tasks, which are strictly sequential, we adopt a milestone-based scoring system. As shown in Table~\ref{tab:task_eval}, these tasks are divided into four logical phases, with intermediate rewards assigned upon the completion of each sub-goal. This stepwise evaluation allows us to pinpoint exactly where a policy might degrade—whether during the initial interaction with deformable objects or during precision-critical phases like zipping or tab insertion.

\begin{table}[t!]
 \centering
  \caption{
  \textbf{Task evaluation standard.} 
  }
  \label{tab:task_eval}
  \footnotesize  
  \begin{tabular*}{\linewidth}{@{\extracolsep{\fill}} c l c l } 
    \toprule
    \textbf{Task} & \textbf{Sub-goals} & \textbf{Total} & \textbf{Score}\\
    \midrule
    \multirow{3}{*}{Conveyor} & Grasp brick & \multirow{3}{*}{10} & 1.0 each \\
     & Place in matched bin &  & 1.5 each \\
     & Workspace cleared &  & 10.0 max \\
     \midrule
    \multirow{4}{*}{Backpack} & Open bag \& Insert items & \multirow{4}{*}{10} & 2.5 \\ 
     & Lift to settle contents &  & 5.0 \\
     & Zip halfway &  & 7.5 \\
     & Zip fully &  & 10.0 max \\
     \midrule
    \multirow{4}{*}{Box} & Load cup & \multirow{4}{*}{10} & 2.5 \\
     & Fold side flaps &  & 5.0 \\
     & Fold rear flap &  & 7.5 \\
     & Tuck locking tab &  & 10.0 max \\
    \bottomrule
  \end{tabular*}
\end{table}

\subsection{Real-World Deployment}
\label{sec:appendix_deploy_details}
To bridge the gap between the discrete, low-frequency inference of the VLA model and the continuous, high-frequency requirements of real-world robotic control, we implemented an asynchronous control framework operating directly in joint space. Specifically, the VLA policy predicts action chunks with a horizon of $H=50$ steps at a relatively low inference frequency, while the robot controller executes joint commands at a 30~Hz frequency. Instead of executing these chunks sequentially, which would cause motion freezing during inference, we adopt a Temporal Ensembling strategy that continuously integrates newly predicted action chunks into a running execution plan.

\begin{figure}[t]
    \centering
    \includegraphics[width=0.98\linewidth]{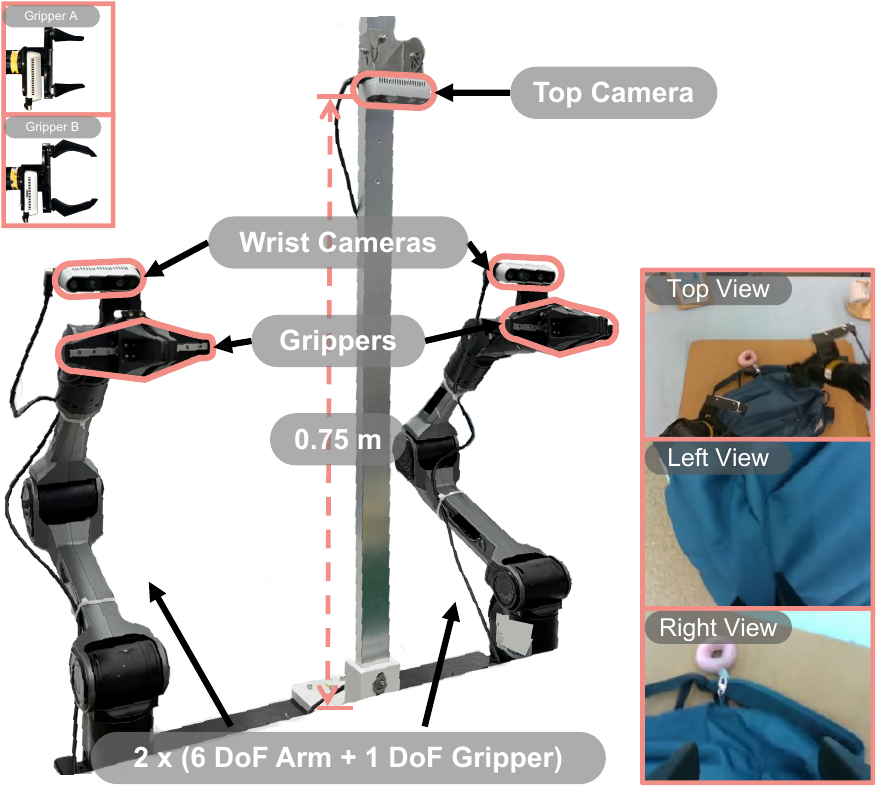}
    \caption{\textbf{Experimental setup.} We utilize a bi-manual platform for our tasks. Each arm possesses 6 DoF along with a 1-DoF gripper, equipped with a wrist-mounted camera. To provide a global view, a top-down camera is positioned centrally between the arms at a height of approximately 0.75 m. The control frequency is set to 30 Hz. 
    \textbf{Top Left:} We apply Gripper A for brick sorting and backpack packing, while applying Gripper B for box closing for the higher precision requirement.
    }
    \vspace{-2mm}
    \label{fig:rl_robot}
\end{figure}

This integration is governed by a linear weighting scheme designed to smooth out transitions and suppress high-frequency jitter. When a new action chunk $\mathbf{a}^{\text{new}}$ is received from the inference thread, it overlaps with the unexecuted portion of the existing action sequence $\mathbf{a}^{\text{old}}$ in the buffer. For any time step $t$ within this overlapping window, the final executed action command $\mathbf{a}_t \in \mathbb{R}^{14}$ (corresponding to the bi-manual setup in Figure~\ref{fig:rl_robot}) is derived via a time-varying linear interpolation between the previous plan $\mathbf{a}^{\text{old}}$ and the new prediction $\mathbf{a}^{\text{new}}$. This ensures that the robot's trajectory is primarily guided by the established plan at the beginning of the update to maintain continuity, while gradually shifting priority to the latest sensory observations towards the end.

\begin{figure*}[ht!]
    \centering
    \includegraphics[width=\linewidth]{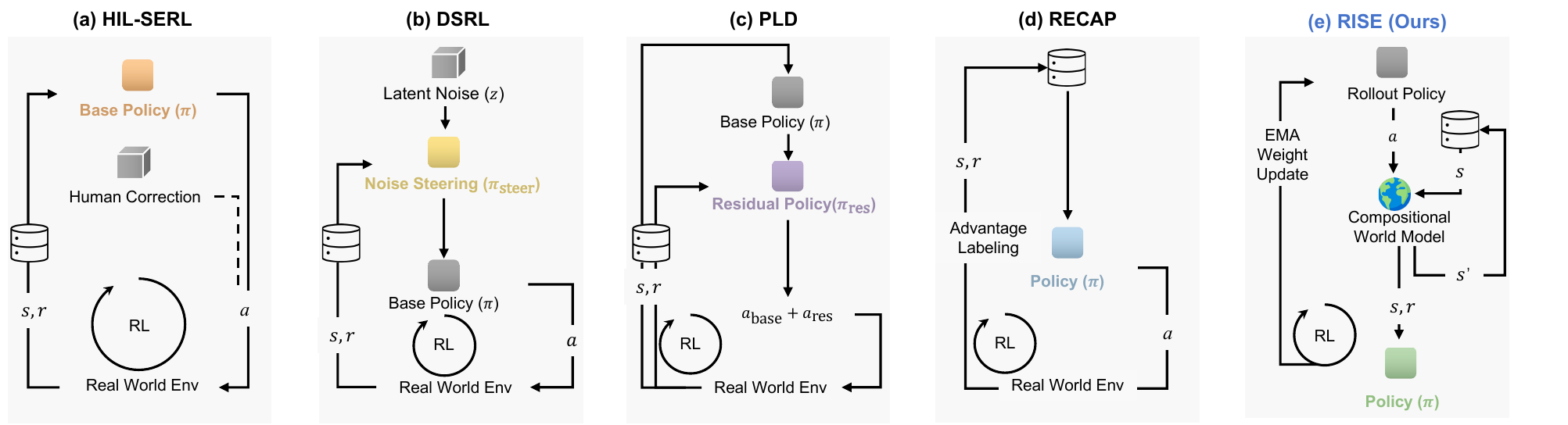}
    \caption{\textbf{Conceptual comparisons with highly-related work.} 
    Different from prior works that heavily rely on off-policy samples 
    from real-world interactions for policy optimization~\cite{luo2025hil-serl,wagenmaker2025dsrl,xiao2025PLD,amin2025pi06},
    \modelname enables on-policy RL by building a world model as an interactive environment.}
    \label{fig:concept}
\end{figure*}

\section{Implementation Details}
\label{sec:implementation_details}

\subsection{Task-specific Data Composition}
Dynamic Brick Sorting includes 3063 human demonstration data and 610 policy rollout data.
Backpack Packing covers 2478 human demonstrations and 507 policy rollout data.
Box Closing features 2286 human demonstrations, 524 policy rollouts, and 540 human corrections (DAgger) data.

\subsection{Dynamics Model}
Our dynamics model operates on multi-view RGB observations ($192 \times 256$) captured from top-down and bilateral wrist cameras, conditioned on future actions. We employ a Flow Matching objective for training. For timestep scheduling, we adopt the Logit-Normal distribution following SD3~\cite{esser2024scaling}, defined as $\operatorname{logit}(t) \sim \mathcal{N}(m, s^2)$, with $m=0.2$ and $s=1.0$. 
Optimization is performed using AdamW with a constant learning rate of $1 \times 10^{-4}$ after a linear warmup for 2k steps. During inference, we solve the flow ODE using the Euler discrete formulation, with 50 denoising steps. 
See \Cref{tab:ours_hyperparams_of_dynamics} for more configurations.

\subsection{Value Model}
The training configurations of the value model are listed in \Cref{tab:config_value_model}.
For each task, the total training takes about 50k steps.
For the first 10k steps, we apply progress estimate loss only,
whereas for the remaining steps, we apply both progress estimate and Temporal Difference learning loss jointly.
Notably, both dynamics model and value model are kept frozen 
during the self-improving loop for policy optimization.

\subsection{Policy Optimization}
\label{sec:pre-training_details}

The policy first gets warmed up mainly following the offline RL approach~\cite{amin2025pi06} with two differences.
RECAP discretizes the labeled advantages into binary bins,
yet we find that discretizing advantages into 10 bins with uniform intervals yields higher results.
Moreover, directly assigning human demonstrations to the highest bins 
while labeling only the policy rollout data
stabilizes learned behavior.
These two discrepancies might emerge from the fact that 
our model initialization $\pi_{0.5}$ is not pre-trained with advantage conditioning, contrary to the offline RL pre-training as in $\pi_{0.6}^{*}$, where RECAP is instantiated.
Subsequently, we start the self-improving loop with configurations listed in~\Cref{tab:ours_hyperparams_self_improve}.

\begin{table}[t!]
    \centering
    \caption{\textbf{Hyper-parameters of dynamics model.}}
    \label{tab:ours_hyperparams_of_dynamics}
    \footnotesize
    \begin{tabular*}{\linewidth}{@{\extracolsep{\fill}} lr @{}} 
        \toprule
        
        \textbf{Hyperparameter} & \textbf{Value} \\
        \midrule
        \textbf{Basics} & \\
        Model initialization & GE-Base~\cite{liao2025genie}\\
        Input / Prediction frames & 4 / 25 \\
        Number of views & 3\\
        Sampling frequency (pre-train / Fine-tune) & 30 / 15 Hz \\
        \midrule
        \textbf{Optimization} & \\
        Training steps (pre-train / Fine-tune) & 120k / 50k \\
        Batch size (pre-train / Fine-tune) & 512 / 64 \\
        Optimizer & AdamW \\
        Learning rate & $1 \times 10^{-4}$ \\
        Conditioned noise level $\sigma$ & 0.2 \\
        \bottomrule
    \end{tabular*}
\end{table}

\begin{table}[t!]
    \centering
    \caption{\textbf{Hyper-parameters of value model.}}
    \label{tab:config_value_model}
    \footnotesize
    \begin{tabular*}{\linewidth}{@{\extracolsep{\fill}} lr @{}} 
        \toprule
        
        \textbf{Hyperparameter} & \textbf{Value} \\
        \midrule
        \textbf{Basics} & \\
        Model initialization & $\pi_{0.5}$~\cite{intelligence2025pi05} \\
        Input frames & 1 \\
        Number of views & 3\\
        \midrule
        \textbf{Optimization} & \\
        Training steps & 50k \\
        Batch size & 64 \\
        Optimizer & AdamW \\
        Learning rate & $2.5 \times 10^{-5}$ \\
        Value discount factor & $0.995$ \\
        \bottomrule
    \end{tabular*}
\end{table}

\begin{table}[t!]
    \centering
    \caption{\textbf{Hyper-parameters of policy self-improving.}}
    \label{tab:ours_hyperparams_self_improve}
    \footnotesize
    \begin{tabular*}{\linewidth}{@{\extracolsep{\fill}} l r } 
        \toprule
        \textbf{Hyperparameter} & \textbf{Value} \\
        \midrule

        \addlinespace[0.3em]
        Batch size & 64 \\
        Optimizer & cosine \\
        Learning rate & $1 \times 10^{-4}$ \\
        Minimum learning rate ratio & 0.1 \\
        Rollout ema decay rate & 0.995 \\
        Action chunk size & 50 \\
        Action dimension & 14 \\
        
        \bottomrule
    \end{tabular*}
\end{table}

\subsection{Baseline Implementation}
\label{sec:appendix_dsrl_details}

Throughout this paper, all policy variants, including baseline and our policy, are instantiated on pre-trained $\pi_{0.5}$ 
to fairly evaluate the effectiveness of various post-training strategies.

\paragraph{$\pi_{0.5}$}
This variant is fine-tuned on our human demonstration corpus only via imitation learning, without using policy rollout or human correction data.

\paragraph{DSRL}
The overall training configurations follow the official implementation of DSRL~\cite{wagenmaker2025dsrl}.
We utilize the $\pi_{0.5}$ model \cite{intelligence2025pi05} as the base policy.
To adapt the policy, we initialize the replay buffer with 10 trajectories collected from the base policy sampled with standard Gaussian noise $w \sim \mathcal{N}(0, I)$, followed by 70 online steering episodes to fine-tune the behavior.

\paragraph{PPO}
We initialize the PPO policy via a pre-trained $\pi_{0.5}$ model. At the rollout stage, we sample real-world trajectories, preserving the inference noise and log probabilities calculated according to RLinf~\cite{zang2025rlinf}. During training, we use this stored inference noise to generate on-policy actions with gradient. We then compute the PPO loss by combining these actions with the new and old log probabilities and advantages. The PPO policy is updated by the PPO loss.

\paragraph{DAgger}
Due to hardware constraints that preclude high-frequency mode switching, we adopt a single-intervention protocol where the human supervisor takes over upon imminent failure and completes the episode.
This variant is trained on both expert demonstrations and additional human correction data via imitation learning.

\paragraph{RECAP}
This variant follows the recipe of the policy warm-up stage, detailed in~\Cref{sec:pre-training_details}.

\begin{figure*}[t!]
    \centering
    \includegraphics[width=1.0\linewidth]{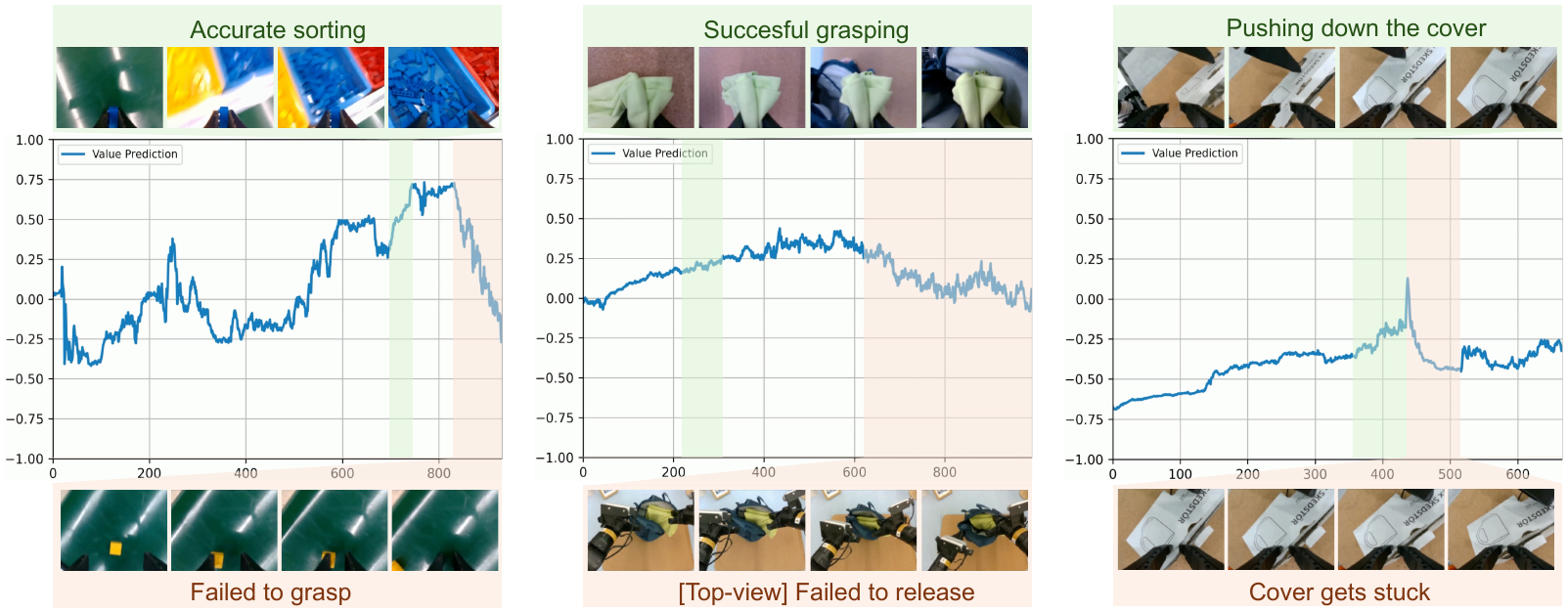}
    \captionof{figure}{\textbf{Qualitative visualizations of value prediction on real-world data.}
    Our value model is capable of distinguishing success and failure,
    highlighted in green and red, respectively.
    }
    \label{fig:value_model}
\end{figure*}

\begin{figure*}[t!]
    \centering
    \includegraphics[width=1.0\linewidth]{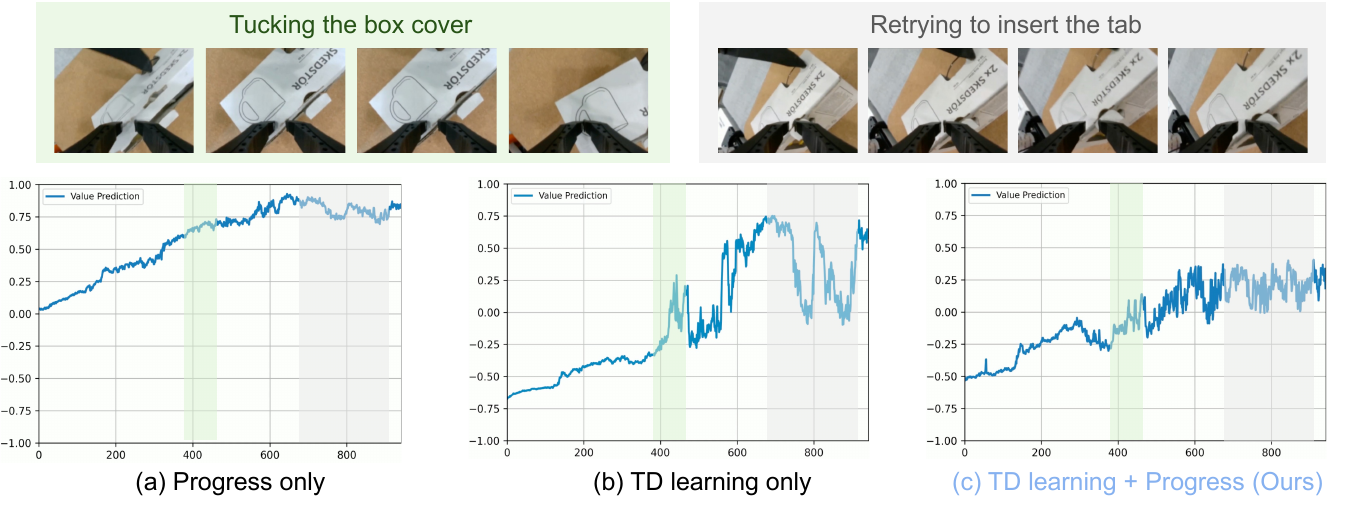}
    \captionof{figure}{\textbf{Qualitative ablation of value model.}
    This visualization ablates the effectiveness of imposing
    each loss during the training of the value model.
    Green and gray regions highlight the favorable and retrying behaviors, respectively.
    In the green region, (b) exhibits a stronger capability in detecting
    critical steps, compared to (a) progress only variant, where the result is simply monotonic.
    However, (b) is less numerically stable compared to (a), as depicted
    in the gray region.
    We jointly apply two losses to feature both visual sensitivity and numerical stability.
    }
    \label{fig:value_ablation}
\end{figure*}

\section{Conceptual Comparisons with Highly-Related Work}
\label{sec:conceptual_comparisons_related_work}

We conceptually compare our method with highly-related work in \Cref{fig:concept}.
Contrary to prior methods that 
learn from off-policy data through costly real-world interactions, 
RISE enables on-policy reinforcement learning with a learned world model that generates new states and assigns advantage for each action chunk.

\section{Qualitative Visualizations}
\label{sec:qualitative_visualizations}

\begin{figure*}[t!]
    \centering
		\includegraphics[width=1.0\textwidth]{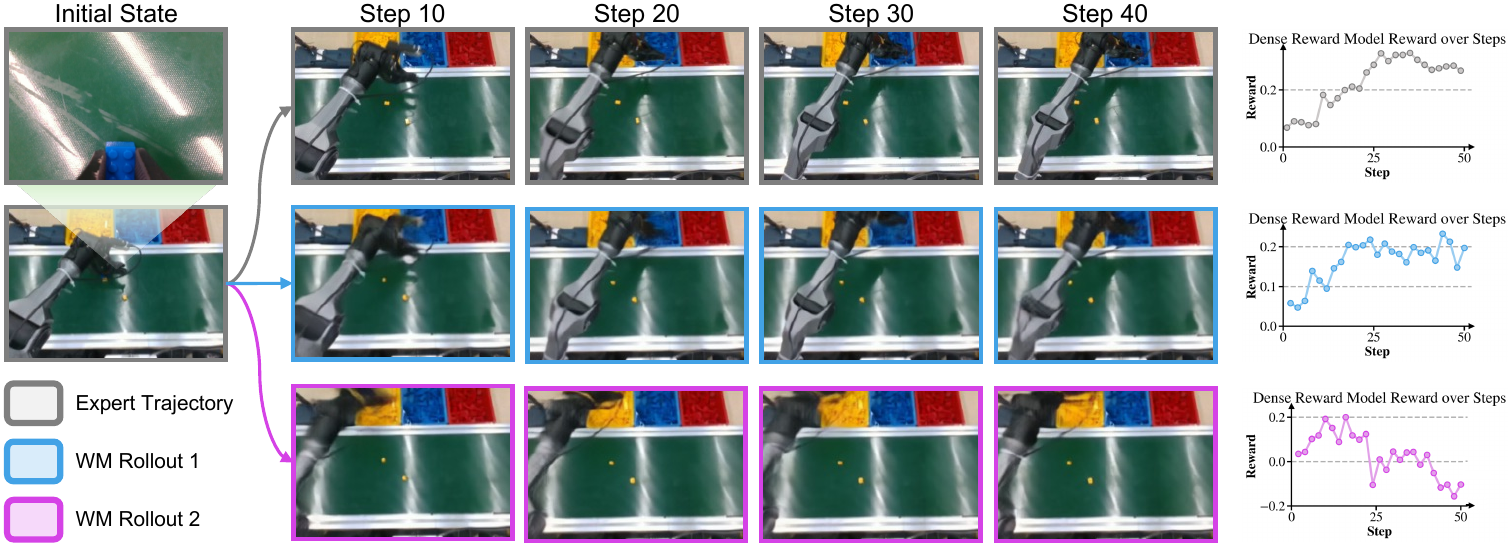}
		\captionof{figure}{\textbf{Multiple rollouts from the same initial state.} 
        \textbf{Left}: Starting from the same state where the gripper grasps a blue brick,
        our world model can synthesize outcomes that accurately follow different actions.
        \textbf{Top Row}: Expert demonstration for reference. 
        \textbf{Middle Row}: Imagined rollout of successful action that 
        correctly put the blue brick into the blue basket, where the rewards go positive.
        \textbf{Bottom Row}: Imagined rollout of failed action that 
        mistakenly put the blue brick into the yellow basket, where the rewards become negative.
        }
  \label{fig:parallel_dream}
\end{figure*}

\noindent \textbf{Compositional World Model.}
We visualize rollout trajectories conditioned on distinct action sequences. As shown in \Cref{fig:parallel_dream}, the dual-arm robot starts with the left arm grasping a blue brick. The expert trajectory executes a smooth pick-and-place operation into the target (blue) box, accompanied by an increasing reward curve. Similarly, the rollout driven by optimized actions exhibits a comparable trend. Notably, the generated video maintains high fidelity, accurately capturing complex environmental dynamics such as the operating conveyor belt. The corresponding reward curve shows improvement but remains slightly below the expert baseline, likely due to minor deviations in the optimized actions or subtle visual artifacts in the imagination. In contrast, the suboptimal trajectory clearly depicts the arm misplacing the brick into the wrong (yellow) box. Consequently, the reward rises during the picking phase but drops significantly once the arm moves toward the incorrect target. These results demonstrate the reliability of our world model in capturing both visual realism and logical consistency.

\begin{figure*}[t!]
    \centering
    \includegraphics[width=1.\linewidth]{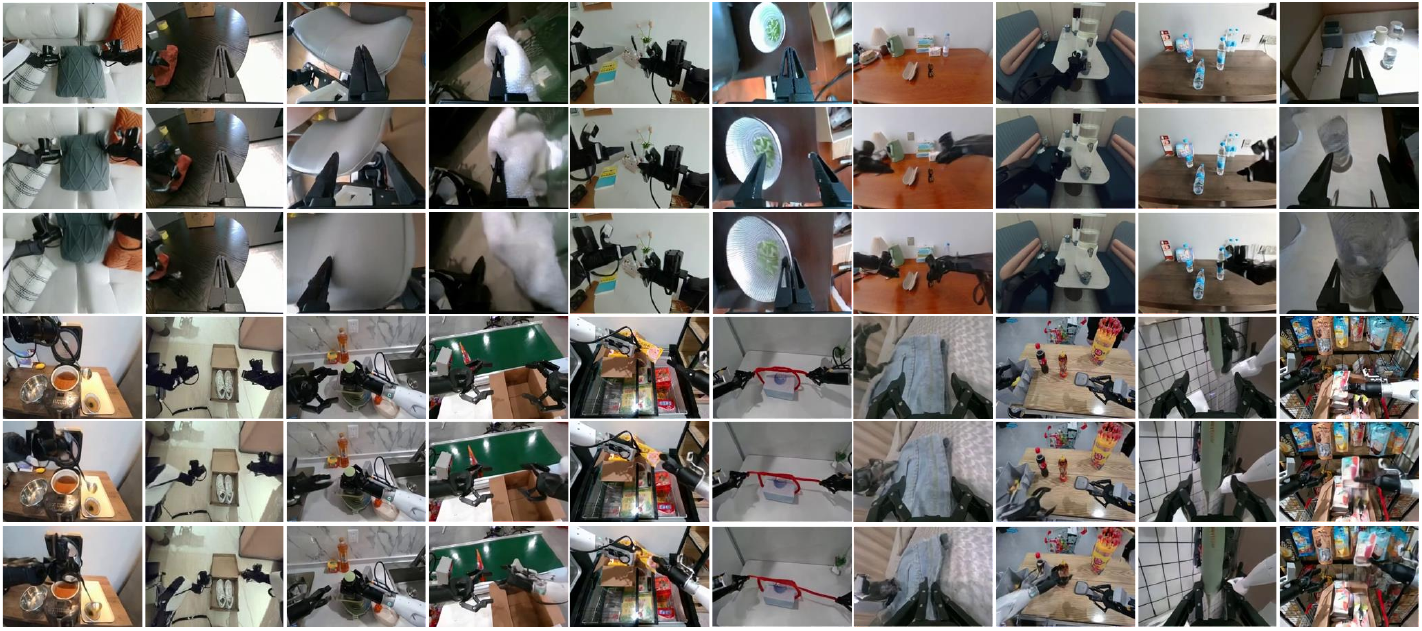}
    \captionof{figure}{\textbf{Dynamics model rollouts.} 
    Each video clip is ordered
    top to bottom.
    }
  \label{fig:dynamics_rollout}
\end{figure*}

\begin{figure*}[t!]
    \centering
    \includegraphics[width=1.\linewidth]{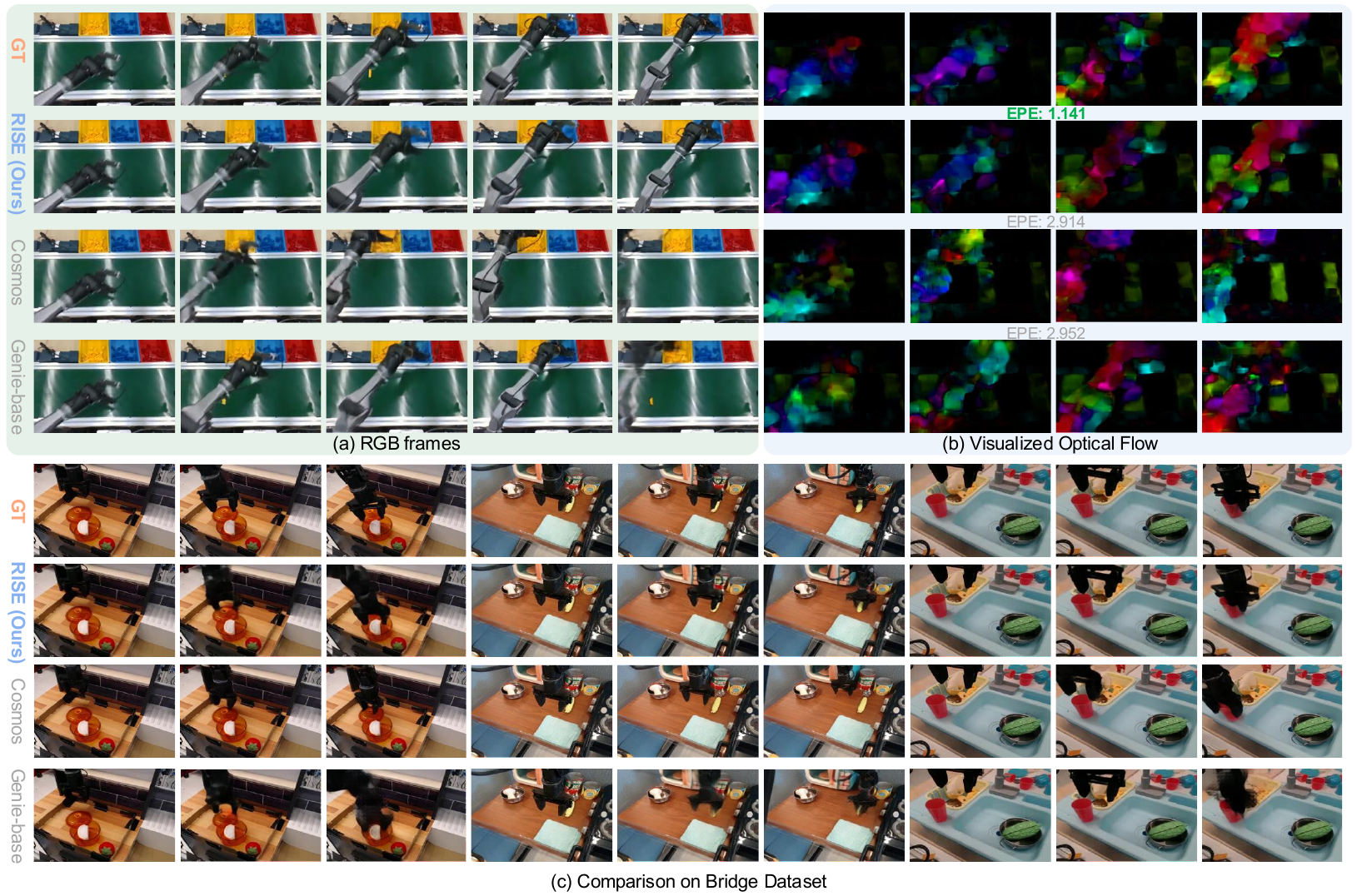}
    \captionof{figure}{\textbf{Comparisons with other generative counterparts.} 
    }
  \label{fig:dynamics_comparison_supp}
\end{figure*}

\noindent \textbf{Dynamics Model.}
We provide a comprehensive visual assessment to benchmark our dynamics model against state-of-the-art alternatives. As shown in ~\Cref{fig:dynamics_comparison_supp}, our method distinctly outperforms other approaches, particularly in maintaining high image quality and precise action alignment. Extending this analysis, \Cref{fig:dynamics_task_multi_view_rollout} and \Cref{fig:dynamics_rollout} present additional multi-view rollouts across real-world tasks, as well as the Galaxea and Agibot World environments, confirming our model's consistency in complex domains.

\noindent \textbf{Value Model.}
We showcase the predicted value trajectory over time alongside corresponding visual observations in~\Cref{fig:value_model}. 
Green regions indicate successful execution, while red regions highlight inferior or suboptimal actions. 
The value model assigns increasing scores during successful executions (e.g., accurate sorting, stable grasping, and successful cover closure), while degrading when subtle failures occur, such as missed grasps, failure to release, or the cover getting stuck. 
Moreover,
we visualize the impact of each loss for training value model in \Cref{fig:value_ablation}.

\section{Failure Modes}
\label{sec:failure_modes}

\begin{figure*}[t!]
    \centering
    \includegraphics[width=1.0\textwidth]{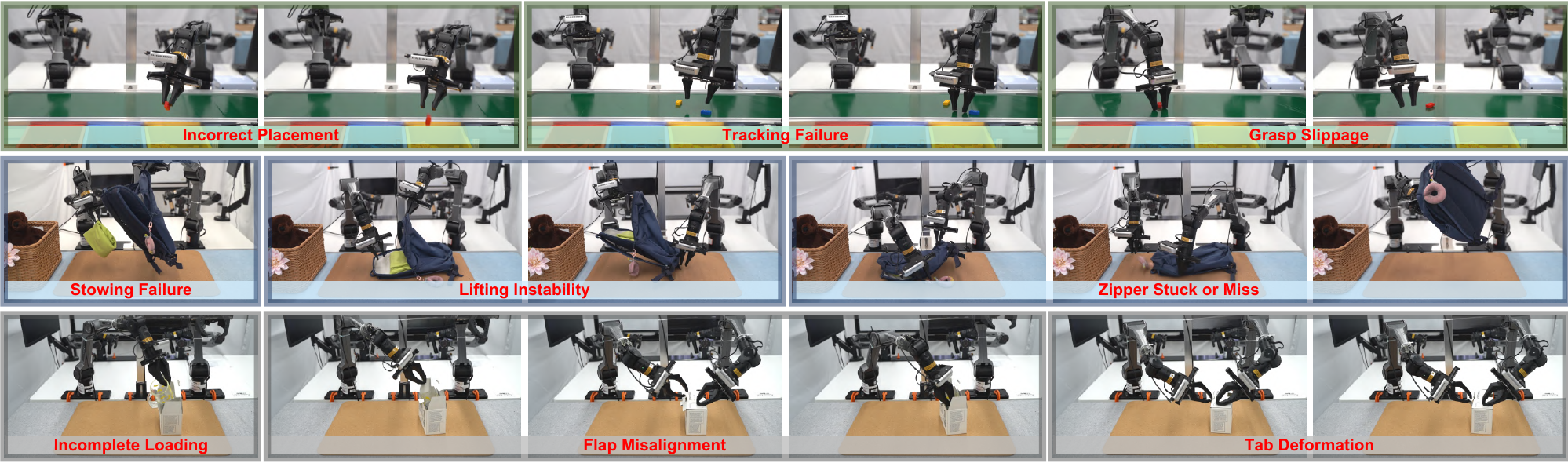}
    \caption{\textbf{Failure modes during inference.} 
    \textbf{\taska{Top:}} Failures typically involve temporal inconsistency in tracking moving objects or precise grasping errors. \textbf{\taskb{Middle:}} The high deformability can lead to incomplete cloth insertion or slippage during the lifting and zipping stages. \textbf{\taskc{Bottom:}} Slight misalignments during bi-manual coordination can cause the cup to tip over during loading or result in unsuccessful folding and tucking.}
    \label{fig:policy_rollout_failure}
\end{figure*}

We depict representative failure behaviors of the \modelname policy in \Cref{fig:policy_rollout_failure}. In \textit{Dynamic Brick Sorting}, failures stem from temporal inconsistency, manifesting as tracking failure or grasp slippage, alongside classification noise. In \textit{Backpack Packing}, high deformability induces stowing failure and lifting instability, while surface compliance leads to zipper stuck or miss. In \textit{Box Closing}, tight geometric tolerances cause incomplete loading, whereas bi-manual synchronization errors result in flap misalignment or tab deformation.

\section{Additional Related Work on VLA Models}
\label{sec:ref_vla}
One recent breakthrough in robot learning is the VLA
framework
that integrates general-purpose vision-language models with low-level robotic control.
Building off pre-trained vision-language models,
RT2~\cite{brohan2023rt2} and OpenVLA~\cite{kim2024openvla} represent actions as discretized bins following the training procedure of language models.
OpenVLA-OFT~\cite{kim2025openvla_oft} parallelizes the decoding process of chunked actions to improve inference latency.
To overcome the multi-modality issue of robot actions
where a variety of actions are correlated with the same state,
GR00T~\cite{bjorck2025gr00t}, $\pi$-series~\cite{black2024pi0, intelligence2025pi05, amin2025pi06}, and RDT~\cite{liu2024rdt} further incorporate action generation with diffusion or flow matching-based architecture inspired by diffusion policy~\cite{chi2023diffusion}.
The massive training of these models is primarily supported by
teleoperated robot datasets~\cite{o2024openx,bu2025agibot, jiang2025galaxea}.
Other data corpora derived from 
simulators~\cite{nasiriany2024robocasa,li2024behavior},
wearable devices~\cite{chi2024umi,wu2026freetacman},
neural synthesis~\cite{jang2025dreamgen}, 
and generic internet~\cite{zhu2023multimodalc4}
are also considered 
for the lack of costly real-world robot data.
Effective approaches are proposed to incorporate heterogeneous 
data sources uniformly, even without sufficient action labels~\cite{univla,ye2024latent,jiang2026wholebodyvla}.
Despite advanced architecture and data scaling,
VLAs still struggle with complex manipulation that 
requires high dexterity and precision~\cite{xiao2025PLD, ghasemipour2025self,amin2025pi06}, where our self-improving approach excels.

\section{License of Assets}
\label{sec:license}

Our dynamics model is built on pre-trained Genie Envisioner~\cite{liao2025genie} under the Apache License 2.0.
The pre-training of our dynamics model leverages two large-scale public datasets, where Agibot World~\cite{bu2025agibot} is under CC BY-NC-SA 4.0 and Galaxea~\cite{jiang2025galaxea} is under Apache-2.0 license.
Some comparisons of the dynamics model are conducted on the Bridge dataset~\cite{walke2023bridgedata} under Creative Commons Attribution 4.0 International License.
Additionally, Cosmos-Predict2.5~\cite{ali2025cosmos2.5} is applied as a baseline under the Apache License 2.0.
Both our policy and value model are initialized from the pre-trained $\pi_{0.5}$~\cite{intelligence2025pi05} under the Apache License 2.0.

\section{Broader Impact}
\label{sec:impact}
Overall, this work contributes to a growing vision of robots that learn continuously and efficiently by reasoning about the consequences of their actions via imagination. 
By improving robustness without excessive physical data collection costs, this work may contribute to safer and more reliable robotic systems that assist humans in physically demanding or hazardous tasks.

\clearpage
\newpage

\end{document}